\documentclass[acmsmall]{acmart}
\usepackage{CJKutf8}
\AtBeginDocument{%
  \providecommand\BibTeX{{%
    \normalfont B\kern-0.5em{\scshape i\kern-0.25em b}\kern-0.8em\TeX}}}

\setcopyright{acmcopyright}%{none}
\copyrightyear{}
\acmYear{2022}
\acmDOI{}

\usepackage{multirow}
\usepackage{lscape}%isn't allowed
\usepackage[figuresright]{rotating}
\usepackage{float}
\usepackage{tabularx}
\usepackage{wrapfig}
\usepackage{amsmath}
\usepackage{graphicx}
\usepackage[caption=false]{subfig}

\usepackage[normalem]{ulem}
\useunder{\uline}{\ul}{}

\newcommand\gpt[1]{\textcolor{black}{#1}}

\newcommand\revision[1]{\textcolor{black}{#1}}

\usepackage{xcolor}
\usepackage{hyperref}
\hypersetup{
    colorlinks=true,
    linkcolor=blue,
    filecolor=magenta,      
    urlcolor=cyan,
    pdfpagemode=FullScreen,
    }

\acmJournal{CSUR}
\acmMonth{2}
\begin{document}
% https://tex.stackexchange.com/questions/365752/how-to-remove-acm-reference-format-box-in-sig-conf-template
\settopmatter{printacmref=false}

\title{Survey of Social Bias in Vision-Language Models}


\author{Nayeon Lee}
\email{nayeon.lee@connect.ust.hk}
\author{Yejin Bang}
\email{yjbang@connect.ust.hk}
\author{Holy Lovenia}
\email{hlovenia@connect.ust.hk}
\author{Samuel Cahyawijaya}
\email{scahyawijaya@connect.ust.hk}
\author{Wenliang Dai}
\email{wdai@connect.ust.hk}
\author{Pascale Fung}
\email{pascale@ece.ust.hk}

%%%%%%%%%%%%%%%%%%%%%%%%%%%%%%%%%%%%%%%%%%%%%%%%
%%%%%%%%%%%%% NAYEON HOLY SAMUEL %%%%%%%%%%%%%%%
%%%%%%%%%%%%%%%%%%%%%%%%%%%%%%%%%%%%%%%%%%%%%%%%
% == New Structure ==
% Bias in Language & Bias in Vision
% Merging Section 3 & 4 into a single "Bias in Unimodal Models" section
% Merge 5.4.1 and 5.4.2 into 5.4
% Focus on Section 5, and revisit the Bias in Language & Bias in Vision later
% Section 5.1 => Move to Section 2
% Section 5.2 => Overview of VL, Keep it
% Section 5.3 => Bridge from Unimodal to VL, Keep it
% Section 5.4 => Break the task into Image2Text and Text2Image (Causes)
% Section 5.5 => Evaluation ()
% Section 5.6 => Mitigation ()
% Section 5.6 => Include in the causes/evaluation of VL
% Section 5.7 => Include in the causes/evaluation of VL

% 5. Bias in VL [Suggestion for NEW structure!!!!]
% 5.1. Unimodal vs multimodal bias (current 5.3)
% 5.2. VL Overview (maybe mention text-to-image and image-to-text)
% 5.3. VL Architectures
% 5.4. VL Bias Causes
% 5.5. VL Bias Evaluation
% 5.6. VL Bias Mitigation

\affiliation{%
  \institution{Center for Artificial Intelligence Research (CAiRE), The Hong Kong University of Science and Technology}
  \city{Clear Water Bay}
  \country{Hong Kong}
%   \postcode{}
}

% \affiliation{%


\renewcommand{\shortauthors}{Nayeon Lee, et al.}

\begin{abstract}
In recent years, the rapid advancement of machine learning (ML) models, particularly transformer-based pre-trained models, has revolutionized Natural Language Processing (NLP) and Computer Vision (CV) fields. However, researchers have discovered that these models can inadvertently capture and reinforce social biases present in their training datasets, leading to potential social harms, such as uneven resource allocation and unfair representation of specific social groups. Addressing these biases and ensuring fairness in artificial intelligence (AI) systems has become a critical concern in the ML community.

The recent introduction of pre-trained vision-and-language (VL) models in the emerging multimodal field demands attention to the potential social biases present in these models as well. Although VL models are susceptible to social bias, there is a limited understanding compared to the extensive discussions on bias in NLP and CV. This survey aims to provide researchers with a high-level insight into the similarities and differences of social bias studies in pre-trained models across NLP, CV, and VL. By examining these perspectives, the survey aims to offer valuable guidelines on how to approach and mitigate social bias in both unimodal and multimodal settings. The findings and recommendations presented here can benefit the ML community, fostering the development of fairer and non-biased AI models in various applications and research endeavors.

% chatGPT version
% In recent years, the rapid progress of machine learning (ML) models, particularly transformer-based pre-trained models, has ushered in a transformative era for Natural Language Processing (NLP) and Computer Vision (CV). Nonetheless, these models have been found to inadvertently inherit and reinforce social biases from their training datasets, potentially leading to social inequities and biased representations of specific social groups. Addressing these concerns and promoting fairness in artificial intelligence (AI) systems has emerged as a paramount focus in the ML community.

% With the advent of pre-trained vision-and-language (VL) models in the burgeoning multimodal field, it becomes crucial to pay attention to the potential social biases present in these models. While VL models are not immune to social bias, the extent of our understanding lags behind the extensive discussions on bias in NLP and CV. This survey seeks to provide researchers with a comprehensive overview of social bias studies across pre-trained models in NLP, CV, and VL domains. By analyzing these perspectives, the survey aims to furnish invaluable guidelines on effectively identifying and mitigating social bias in both unimodal and multimodal settings. The insights and recommendations garnered from this study can significantly benefit the ML community, fostering the development of fairer and unbiased AI models across various applications and research endeavors.

% \textbf{Abstract - TODO}
\end{abstract}

% %%
% %% The code below is generated by the tool at .
% % https://dl.acm.org/ccs
% %% Please copy and paste the code instead of the example below.
% %   <concept>
% %       <concept_id>10002944.10011122.10002945</concept_id>
% %       <concept_desc>General and reference~Surveys and overviews</concept_desc>
% %       <concept_significance>500</concept_significance>
% %       </concept>
%   <concept>
%       <concept_id>10010147.10010178.10010179.10010182</concept_id>
%       <concept_desc>Computing methodologies~Natural language generation</concept_desc>
%       <concept_significance>500</concept_significance>
%       </concept>
%   <concept>
%       <concept_id>10010147.10010257.10010293.10010294</concept_id>
%       <concept_desc>Computing methodologies~Neural networks</concept_desc>
%       <concept_significance>500</concept_significance>
%       </concept>
%  </ccs2012>

% \ccsdesc[500]{General and reference~Surveys and overviews}
% \ccsdesc[500]{Computing methodologies~Natural language generation}
% \ccsdesc[500]{Computing methodologies~Neural networks}

\maketitle

\tableofcontents
% \newpage

\section{Introduction}
\label{section:introduction}
% Limit this section to 1 page

Recently, researchers discover that machine learning (ML) models have the ability to capture and learn 
% the discriminating social biases 
\gpt{from the social biases that can lead to discrimination}
that exist in the training dataset. Such social bias in ML systems can 
% cause social harms such as uneven allocation of resources or opportunities (allocational harm) and/or unfavorable representation, and failure of recognition, or demand of particular social groups (representational harm) 
\gpt{cause social harms, such as the uneven allocation of resources or opportunities (allocational harm) and the unfavorable representation, failure of recognition, or overlooked demands of particular social groups (representational harm)}
\cite{barocas2017problem,blodgett-etal-2020-language}.
The International Covenant on Civil and Political Rights (The United Nations General Assembly 1966) encodes that ``the law shall prohibit any discrimination and guarantee to all persons equal and effective protection against discrimination on any ground such as race, color, sex, language, religion, political or other opinion, national or social origin, property, birth or other status.''. Various international bodies and individual governments have also issued AI ethics guidelines that include ``fairness'' as an important criterion for AI and ML models and systems \cite{floridi2021ethical, OECD_2019, European_commision_2019, jobin2019global, Vought_2021}.
% \pascale{can you cite some important AI guidelines also?} \yeon{@bang} 
Thus, ensuring fairness in artificial intelligence (AI) systems by mitigating such social biases has become an increasingly important 
% issue 
\gpt{focus}
in the ML community.

In the past few years, Natural Language Processing (NLP) and Computer Vision (CV) have experienced huge leaps thanks to transformer-based pre-trained models. As such pre-trained models become widely adopted in each of the fields for research and applications, 
% fairer and non-biased models are in dire. 
\gpt{there is a dire need for fairer and non-biased models.}
As a response to the urgent call, there are many studies on social bias in NLP and CV, but they remain separate. Moreover, with the advancement of the multimodal field, especially the Diffusion Model in vision-and-language (VL), such fairness investigations in pre-trained VL models are also conducted. These VL models are indeed susceptible to social bias, but limited understanding has been shared compared to in-depth discussions of social bias in NLP or CV. This survey provides researchers with a high-level 
% \pascale{why do you say it is high level?} 
insight derived 
% from the similarities and differences of social bias studies from NLP, CV, and VL, so it can benefit the community by providing guidelines on how to approach the problem of social bias in both unimodal and multimodal settings.
\gpt{from the similarities and differences in social bias studies across NLP, CV, and VL. It aims to benefit the community by providing guidelines for addressing social bias in both unimodal and multimodal settings.}
% in the vision and language models. 
% Moreover, we provide current progress on social bias in VL models. 
% In the fields of NLP and CV, there have been in-depth discussions on bias metrics, mitigation, and analysis, of which insights could be shared. 

% bias -- some are desirable, and some are not desirable. 

% \yeon{@nayeon TODO: add some positive things about our advancement in bias mitigation. And let's try to not just criticize problematic behaviors of machine learning models.}

\paragraph{Organization of this Survey.} The remainder of this survey is organized as follows. Section~\ref{sec:overview} illustrates the overview regarding the concept of bias in machine learning as well as the common metrics and general frameworks for mitigating bias in machine learning. Section~\ref{sec:bias-unimodal} presents the causes, evaluation, and mitigation methods of social bias in the case of modeling a single modality, covering bias in the textual (language) modality and bias in the vision modality. Section~\ref{sec:bias-vl} highlights the works on bias in vision-and-language multimodality, as well as emphasizes the causes, evaluation, and mitigation techniques in VL models.

\section{Overview}
% Limit this section to 3 pages
\label{sec:overview}

\subsection{Concept and Terminology}
% Every machine learning model is designed to learn a certain form of model \pascale{statistical} bias, which helps it to make the correct decision for the intended task.
\revision{Machine learning models are designed to learn a specific form of model statistical bias called inductive bias, which helps them make accurate decisions for their intended task \cite{saunshi2022understanding, hellstrom2020bias, goyal2022inductive, cadene2019rubi, battaglia2018relational}. Inductive bias refers to the inherent assumptions or prior knowledge that models incorporate during learning, guiding them to favor certain solutions based on data patterns. However, models can unintentionally learn 
\textbf{social bias}, leading to unfair favoritism or discrimination.}
% \pascale{This is not clear. You need to have a lot of citations.}
% Our focus is not on such inductive bias, but an unintended social bias that the models learn.
\textbf{Social bias} is defined as the disproportionate weight in favor of or against one thing, person, or group compared with another, usually in a way considered to be unfair. It is ``unfair'' if individuals are not treated equally, especially in terms of opportunity, based on {protected} attributes~\cite{binns2018fairness, kusner2017counterfactual, binns2020apparent, jiang2020identifying, garrido2021survey}.
\textbf{\textit{Protected} attributes} refer to sensitive demographic factors 
% that the model should stay independent 
\gpt{from which the model should remain independent}
(e.g., gender, race, age, marital status, etc.). 
% A system should be stereotype-free, where a stereotype is a preconceived idea that certain attributes or characteristics apply to all members of a specific demographic group. The existence of stereotypes indicates the existence of biased behavior that is based on preconceived beliefs against a certain demographic group. Some examples of stereotypes are gender stereotypes in occupations (male=doctor, female=nurse) and race stereotypes (black=higher chance of committing a crime).

% One close line of works is about ``robustness''. Fairness can be considered as a part of or closely related concept of ``Robustness under Distribution Shift'' as categorized by \cite{wang-etal-2022-measure}. In this paper, we focus on ``social bias'' under the umbrella of the research topic of fairness.

\subsection{Protected Demographic Attributes}
% \subsection{Social Bias Definition / Demographic Categorization}
% \pascale{include other attributes beyond race/gender}
\revision{
Many nations prohibit the discrimination \gpt{against} individuals based on the following \textit{protected attributes} of people~\cite{protected_group_wiki}: sex (including sexual orientation and gender identity), race (including color and ethnicity), age, religion, disability, national origin, creed, marital status, pregnancy, family status, and genetic information.
In the context of AI research, gender and race are the two most heavily explored and mitigated attributes, motivated by controversial algorithmic errors that gained public attention such as i) \gpt{an} image-recognition algorithm auto-tagged pictures of black people as ``gorillas''~\cite{BBC_News_2015}, ii) \gpt{an} image retrieval result of ``CEO'' favoring male images~\cite{UW_News_2015}, iii) \gpt{a} chatbot generating toxic and racist response\gpt{s}~\cite{Wikipedia_2023b}.
This survey also focuses on gender and racial bias, but we would like to emphasize that other protected attributes are as important as race/gender and require attention from researchers. We believe the insights obtained from racial and gender bias work can serve as good guidance for facilitating such future work.}
% some attribute information (e.g., race, gender) is more conspicuous and recognizable than others (e.g., national origin, marital status, family status, political orientation). 

% One of the most important considerations in racial and gender bias work is the categorization of subgroups within race and gender. Here, we describe these categorizations adopted in both CV and NLP fields as a preliminary understanding for readers. 
\gpt{An essential consideration in racial and gender bias work is the categorization of subgroups within race and gender. In the following, we outline these categorizations as they are adopted in both the CV and NLP fields to provide readers with a foundational understanding.}

\subsubsection{Gender/Sex Subgroups}
Gender is a socially constructed characteristic that can change over time and varies from society to society,\footnote{
\url{https://www.who.int/health-topics/gender}} meaning there is no fixed number of categories. Gender is often interchangeably confused with sex, which refers to biological and physiological characteristics (e.g., female and male), although they are distinct. For the simplicity of exploration, gender is often categorized in a simple binary manner (i.e., man, woman) in much of the literature in both NLP
~\cite{bolukbasi2016man,rudinger-etal-2018-gender,sheng2019woman,kaneko-bollegala-2019-gender,zhao-etal-2019-gender,maudslay2019s} and Vision~\cite{wang2019balanced,bhargava2019exposing}. Along with the increased LGBTQ+ awareness, efforts to construct inclusive AI research that goes beyond binary gender have been made; for instance, \citet{hazirbas2022casual} suggest categories of ``Cis Man, Cis Woman, Transgender Man, Transgender Woman, Non-binary, Not stated above'' to be inclusive.

%%% challenge
% There are distinct challenges in categorizing gender in both NLP and CV. 
\gpt{Categorizing gender in both NLP and CV presents unique challenges. For NLP, }
% In the case of NLP, 
many crowdsourced data or web data often lack self-identification, or the study leverages data with perceived genders (e.g., ``He is xyz.''). Also, there are no clear pronouns for all genders besides ``she'' and ``he'', so ``they'' is increasingly being used to refer to non-binary/undefined genders in parallel with its traditional meaning of ``two or more people''. However, many NLP bias works leverage traditional pronouns and gender words\footnote{he, she, male, female, man, woman, men, women, gentlemen, lady, boy, girl, mom, dad, etc.} to collect data or test the disparities in different groups. In CV, unless the participants of the collected dataset disclose or annotate themselves, it is impossible to avoid annotators' bias.
% \yeon{( @nayeon TODO elaborate on this)}.
For both fields, the genders beyond binary classification are much needed to avoid bias and encourage inclusive AI research. 
% % Despite these efforts having laid down the groundwork, some challenges still persist in inclusive AI research.

% However, it is still on-going effort to devise a way for it. 

\subsubsection{Race/Ethnicity Subgroups}
In NLP, it is common to explore bias with a pre-defined list of races, ranging from (Black, White) \cite{sheng2019woman} to a more expanded list such as (African, American, Asian, Black, Hispanic, Indian, Pacific Islander, White) \cite{sheng2021revealing}.

\begin{figure}[t]
    \centering
    \includegraphics[width=0.8\linewidth]{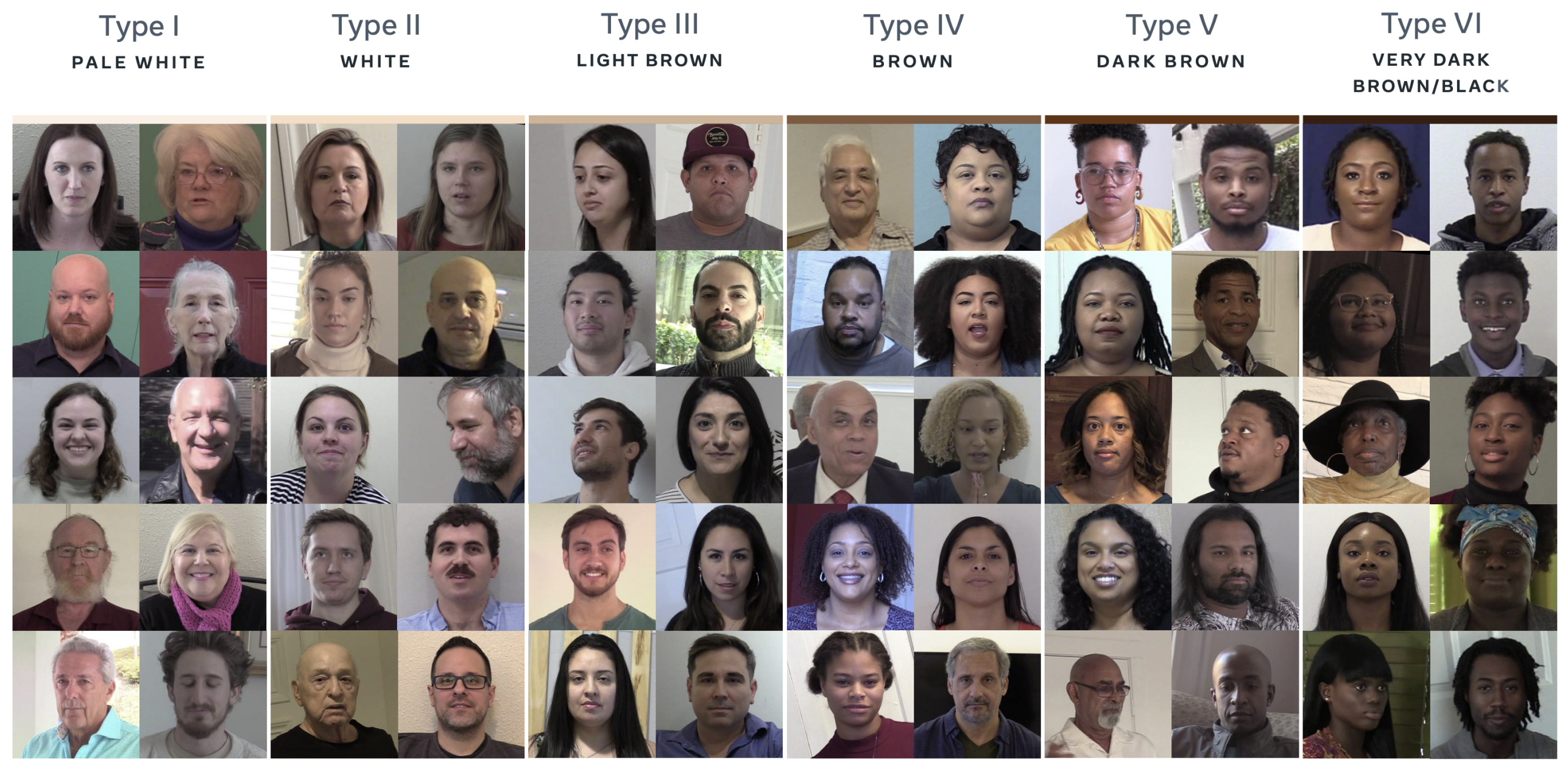}
    \caption{Illustration of the Fitzpatrick scale~\cite{fitzpatrick1975soleil} from \citet{hazirbas2021towards}.}
    \label{fig:fitzpatrick}
\end{figure}

However, categorizing racial and ethnic subgroups in CV can be more challenging than in NLP and more prone to controversies \cite{roth2016multiple}.
The most common two categorization methods in CV are the following.
% The first method is, as same as NLP, to have predefined race lists. 
\gpt{The first method is similar to that of NLP, which uses predefined race lists.}
% Different works have different lists of race, 
\gpt{Different studies utilize varying lists of races,}
ranging from binary (Black, White)~\cite{buolamwini2018gender} to septenary (Western White, Middle Eastern, East Asian, Southeast Asian, Black, Indian, Latino)~\cite{karkkainen2021fairface}. 
% \yeon{(Also, there is general concern of the validity of man-defined subgroups -- for both nlp and vision) } 
The second method, which is unique to CV, is to leverage skin tones such as the Fitzpatrick scale~\cite{fitzpatrick1975soleil} or the Monk Skin Tone (MST) Scale~\cite{monk2014skin}. The Fitzpatrick scale~\cite{fitzpatrick1975soleil} is a numerical classification schema for human skin color developed to estimate the response of different skin types to ultraviolet (UV) light. Fig.~\ref{fig:fitzpatrick} shows examples of face categorization based on the Fitzpatrick scale. Some works use the full 6-scale of the Fitzpatrick scale, but some works map 1-3 into the ``lighter skin tone'' class and 4-6 to the ``darker skin tone'' class for a binary setup. The MST scale has an extended scale with 10 different skin tones, in which the darker and the lighter shades are distributed equally.

\citet{hazirbas2021towards, hazirbas2022casual} provide a very detailed discussion about defining ``race/ethnicity'' in vision using skin tone (i.e., Fitzpatrick scale). The first point is that race is a social construct, which is fluid and has ambiguities that make it hard to define in a universally agreeable way \cite{roth2016multiple, hanna2020towards}, therefore, skin tone has been adopted as a ``expressive and generic way to group people''\cite{hazirbas2021towards}. The second point is about the reliability of the most commonly used scale Fitzpatrick scale, which relies on Western theories and performs worse on darker skin tones \cite{ware2020racial,okoji2021equity}. Thus, \citet{hazirbas2022casual} suggests the use of both Fitzpatrick and MST scales for annotation so that Fitzpatrick can be used for comparison with previous works while MST can ensure a more balanced representation.

\subsection{Fairness Criterion}
\label{subsec:ml_bias}
% \pascale{ motivate and tell why these papers? }
The first step of understanding bias is to define a way to measure and quantify bias. 
Since the concept of bias is closely associated with ``fairness'', it is important to define the fairness criteria. Many \gpt{studies} utilize fairness criteria to formulate the metric for bias measurement.

Given that $Y$ is the target or true outcome, $Z$ is \textit{protected attribute} and $\hat{Y}$ is the model prediction ($\hat{Y} \in \{0, 1\}, Z \in \{0, 1\}$), the foundational fairness criteria are the following:

\begin{itemize}
\item [$\bullet$] {\bf Demographic Parity} \cite{zemel2013learning}: A decision (e.g., accepting or denying a loan application) must be independent of the \textit{protected} attribute (e.g., race, gender). In other words, it requires the models to have the same positive rate among all demographic groups.
\begin{align*}
    & Pr\{\hat{Y} = 1 | Z = 0\} = Pr\{\hat{Y} = 1 | Z = 1\}
\end{align*}
Although demographic parity appears to be a good constraint for fairness, \cite{hardt2016equality} pointed out its weakness: it does not always fully ensure fairness and may cripple machine learning. To ensure all demographic groups have equal positive rates, demographic parity allows unqualified individuals (i.e., $Y=0$) to be accepted (i.e., classified as $Y=1$). Giving opportunities to unqualified individuals over qualified individuals cannot be considered "fair". Also, if the target variable $Y$ is highly correlated with \textit{protected} variable, demographic parity will not allow the model prediction to ever be optimal. 

\item [$\bullet$] {\bf Equality of Odds} \cite{hardt2016equality}: A decision must be independent of the \textit{protected} attribute conditional on $Y$. For positive outcome $y=1$, the constraint requires that $Y$ has equal \textit{true positive rates} across the two demographics $Z = 0$ and $Z = 1$. For negative outcome $y=0$, the constraint equalizes false positive rates. 
\begin{align*}
    & Pr\{\hat{Y} = 1 | Z = 0, Y = y\} = Pr\{\hat{Y} = 1 | Z = 1, Y = y\}, y \in \{0, 1\}
\end{align*}
Equality of odds is the improvement of demographic parity because it aligns nicely with the central goal of building highly accurate classifiers. It does not restrict the optimal model in the learning process; it simply enforces the accuracy to be equally high for all demographics (i.e., punishes models that perform well only on the majority).

\item [$\bullet$] {\bf Equality of Opportunity} \cite{hardt2016equality}: A decision must be independent of the \textit{protected} attribute conditional on $Y$, but only for the case $Y=1$. It is a relaxed version of ``equalized odds'' that only requires non-discrimination within the``advantaged'' target group ($Y=1$). In other words, it requires the models to have the same true positive rate among all demographic groups.
\begin{align*}
    & Pr\{\hat{Y} = 1 | Z = 0, Y = 1\} = Pr\{\hat{Y} = 1 | Z = 1, Y = 1\}
\end{align*}

\end{itemize}

\subsection{Categorization of Bias Metrics and Mitigation Methods}
\subsubsection{Bias Metrics}
\revision{There exist several techniques to measure social bias. Inspired by \cite{goldfarb2021intrinsic,delobelle2022measuring}, we categorize the bias measurement techniques into two \gpt{main} categories, \textit{Intrinsic} and \textit{Extrinsic}, throughout this paper.}
% For ease of understanding, we provide an overview by categorizing the bias measurement techniques into \textit{Intrinsic} and \textit{Extrinsic}, inspired by \cite{goldfarb2021intrinsic, delobelle2022measuring}, }.
\begin{itemize}
    \item \textit{\textbf{Intrinsic}} bias metric measures the bias that exists in learned embedding/feature space of pre-trained models.
    \item \textit{\textbf{Extrinsic}} bias metrics measure bias through extrinsic downstream tasks (e.g., text classification, dialogue generation, image captioning, image classification, etc.) by measuring the disparity in performance among different social groups.
\end{itemize}

\revision{
It is crucial to mitigate extrinsic bias as it is associated with model error and can directly impact the end-users. For instance, auto-tagging black people as ``gorillas'' on photo albums will exert explicit harm and discrimination towards end-users, and thus, should strongly be avoided. On the contrary, there are conflicting views regarding the mitigation of intrinsic bias for the following reasons:
1) Some argue that bias in embedding space will not directly impact the end users, and many research works have discovered that bias in embedding space does not always propagate into downstream tasks~\cite{goldfarb2021intrinsic, delobelle2022measuring}. Yet, some argue that there is no 100\% guarantee that embedding-level bias will truly not propagate down---since deep learning models are similar to a black box, we cannot be confident about anything.
% Moreover, intrinsic bias does not always mean bias propagated in downstream tasks (i.e., extrinsic bias) \cite{goldfarb2021intrinsic}
2) Some argue that some learned gender/race features are not harmful and are necessary for the models to perform well. As a result, mitigation of intrinsic bias often leads to unwanted performance degradation. 
% \yeon{(TODO: we need more citations)}
}

\subsubsection{Mitigation Methods}
% There exist common ML fairness mitigation methods that are adopted to both vision and language. We describe them more in details how they are adopted in later sections (section references). Moreover, due to the different natures of the fields, there also exist specific mitigation methods (e.g., word embedding, etc). 
% There exists three family of mitigation methods adopted in both the CV and the NLP field. We categorize our review of various mitigation methods following this categorization:
As summarized in \cite{mehrabi2021survey}, there are three primary families of mitigation methods that are prevalent in both the CV and the NLP domains:
\begin{itemize}
    \item 1) \textbf{Pre-processing} techniques are applied to the data to remove the bias and discrimination inherent in the data.
    
    \item 2) \textbf{In-processing} techniques modify the learning algorithms with the goal of mitigating discrimination during the model training phase. In-processing techniques are normally in the form of changing the optimization objective or imposing constraints.
    
    \item 3) \textbf{Post-processing} techniques are applied during the inference phase to modify and fix the model behavior after training. This method is especially helpful when the model is only available in a black box manner (i.e., without any ability to modify the training data or learning algorithm). Post-processing techniques reassigned the labels predicted by the black-box model based on a function.
\end{itemize}

\section{Bias in Unimodal Models}
% Limit this section to 7 pages

\label{sec:bias-unimodal}

\revision{
% chatGPT
Multimodal challenges in machine learning demand the integration of visual and textual information, wherein VL models play a pivotal role. 
% To tackle multimodal challenges, VL models play a crucial role in integrating information from both visual and textual modalities. 
However, an inherent concern lies in the presence of social bias during the unimodal pre-training stages that these models undergo~\cite{cadene2019rubi, mbintvqa, abbasnejad2020counterfactual, booth2021bias, zhang2022counterfactually}. This bias becomes ingrained and carried forward, potentially perpetuating harmful stereotypes and exacerbating societal disparities. Addressing this issue necessitates a focus on refining these models to mitigate bias and promote fairness.}

% Multimodal challenges in machine learning demand the integration of visual and textual information, wherein Visual-Linguistic (VL) models play a pivotal role. However, a significant concern emerges from the presence of social bias during the unimodal pre-training stages of these models, as evidenced in previous studies~\cite{cadene2019rubi, mbintvqa, abbasnejad2020counterfactual, booth2021bias, zhang2022counterfactually}. This bias tends to become deeply ingrained, perpetuating harmful stereotypes and exacerbating societal disparities. To address this pressing issue, our paper focuses on refining VL models to effectively mitigate bias and foster fairness, ultimately leading to more equitable AI solutions.
% \pascale{you need to state the motivation for this section 3. Why do people want to read about text only or vision only models here and are these long enough? If not you need to state why this is only 7 pages.}

\subsection{Bias in Language}
% Social bias and fairness have 

\subsubsection{Measuring and Analyzing Bias}
%이런식으로 쓰고싶음~ >_< \bang{In [17], authors studied fairness definitions in political philosophy and tried to tie them to machine learning. Authors in [70] studied the 50-year history of fairness definitions in the areas of education and machine learning. In [149], authors listed and explained some of the definitions used for fairness in algorithmic classification problems. In [139], authors studied the general public’s perception of some of these fairness definitions in computer science literature. Here we will reiterate and provide some of the most widely used definitions, along with their explanations inspired from [149].}
\label{subsec:nlp_bias}
There are various bias measurement techniques that quantify social bias in NLP, including commonly known techniques such as Word Embedding Association Test (WEAT)\cite{caliskan2017semantics}, Sentence Encoder Association Test (SEAT) \cite{SEAT}, \textit{iCAT} (a.k.a. StereoSet) \cite{nadeem2021stereoset}, and Crowdsourced Stereotype Pairs (CrowS-Pairs) \cite{nangia2020crows}.
Additionally, there are different ways of categorizing the measuring techniques. \citet{sun2019mitigating} provide an overview of measurements of gender bias in NLP. \citet{czarnowska2021quantifying} categorize the existing 22 extrinsic fairness metrics in NLP into their three generalized fairness metrics. \citet{delobelle2022measuring} study social bias metrics for pre-trained language models with a metric categorization of intrinsic and extrinsic.
\textit{Intrinsic} bias metrics measure the bias that exists in word embedding \cite{goldfarb2021intrinsic} or the bias in language model from pre-training \cite{delobelle2022measuring}. \textit{Extrinsic} bias metrics measure bias through extrinsic downstream tasks by measuring the disparity in performance among different social groups.

% For ease of understanding, we provide an overview by categorizing the bias measurement techniques into \textit{Intrinsic} and \textit{Extrinsic}, inspired by \cite{goldfarb2021intrinsic, delobelle2022measuring}. 

\paragraph{\textbf{Intrinsic Bias Measure}}
\label{subsubsec:nlp_intrinsic_bias}

Intrinsic bias can be understood as any social bias captured within word-embedding or language model itself before it is applied to any downstream tasks.
% === word embedding (WEAT) === 
The most common intrinsic bias metrics are the Word-Embedding Association Test (WEAT) \cite{caliskan2017semantics} and its variants: Sentence-Embedding Association Test (SEAT) \cite{SEAT}, \citet{tan2019assessing}, and Contextualized Embedding Association Test (CEAT) \cite{guo2021detecting}. WEAT is inspired by a psychological test that quantitively measures humans' subconscious gender bias through the difference in time and accuracy when categorizing words into two different concepts, which is called Implicit Association Test (IAT) \cite{greenwald1998measuring}. Similarly, WEAT measures bias by comparing cosine similarities among two sets of target words (e.g., $X$ = \{programmer, engineer, ...\} $Y$=\{nurse, teacher, ...\}) and two sets of attribute words (e.g., $A$ = \{man, male\} and $B$ = \{woman, female\}) as follows:
\begin{center}
$s(X,Y,A,B) = \sum_{x\in~X}s(x,A,B) - \sum_{y\in~Y}s(y,A,B)$
\end{center}
, where $s(w,A,B)$ is defined as $mean_{a\in{A}} cos(w, a) -mean_{b\in{B}} cos(w, B)$.
\citet{caliskan2017semantics} include gender, racial, and age-group bias in target groups (e.g., European-American vs. African-American names; young vs. old people's names) with pleasant and unpleasant attribute words.

\citet{bolukbasi2016man,manzini2019black} measure gender, racial, and religious bias through an analogy task that calculates differences in distances from a neutral word vector to the vectors of a pair of bias-specific words (i.e., $\overrightarrow{doctor}$ is closer to $\overrightarrow{man}$ than to $\overrightarrow{woman}$). An example of implicit bias in word embedding through an analogy puzzle is as follows: ``a man is to computer programmer as a woman is to x'' (denoted as $man:computer programmer = woman:x$), simple arithmetic of the embedding vectors finds that $x = homemaker$ is the best answer because:
\begin{center}
$\overrightarrow{man} - \overrightarrow{woman} \approx \overrightarrow{computer programmer} - \overrightarrow{homemaker}$
\end{center}, which elicits gender bias in the two distinct occupations.

Another common bias metric, Context Association Test (CAT) \cite{nadeem2021stereoset}, measures stereotypical bias in gender, profession, race, and religion in pre-trained language models (LMs) (Stereotype Score; \textit{ss}) as well as its language model ability (Language Modeling Score; \textit{lms}), which is converted into a single metric Idealized CAT Score (\textit{icat}). \textit{ss} measures how often the LM prefers stereotypical association over an anti-stereotypical association while \textit{lms} measures the frequency of the LM prefers the meaningful over meaningless association. CAT consists of both a sentence-level test (\textit{intrasentence}) and a discourse-level test (\textit{intersentence}). This test is often referred to as StereoSet. 
% The ideal score should be 100. They tested on BERT, RoBERTa, XLNET, GPT-2. the best model is still 27.1 behind. large streotype in it. 

\citet{nangia2020crows} measure the percentage of samples where a language model assigns a higher pseudo-loglikelihood \cite{salazar-etal-2020-masked} to the stereotyping sentence over the less stereotyping sentence given sensitive/protected attribute-related tokens (e.g., he, she), with its dataset called Crowdsourced Stereotype Pairs (CrowS-Pairs). The dataset covers a broad range of sensitive attributes: race, gender, sexual orientation, religion, age, nationality, disability, physical appearance, and socioeconomic status. Each pair of the dataset consists of minimally distant sentences, one is stereotypical and another is anti-stereotypical. 
% In detail, this metric specifically adopts pseudo-likelihood to mimic perplexity-based scoring for masked language model over a sentence given sensitive/protected attribute-related tokens (e.g., he, she). 
\citet{neveol-etal-2022-french} further extend the CrowS-Pairs to French. 

% \citet{nangia2020crows} measures the percentage of examples where a language model assigns a higher psuedo-loglikelihood \cite{salazar-etal-2020-masked} to the stereotyping sentence over the less stereotyping sentence, with its dataset called Crowdsourced Stereotype Pairs (CrowS-Pairs).  Each pair of the dataset consists of minimally distant sentences, one is stereotypical and another is anti-stereotypical. In detail, this metric specifically adopts pseudo-likelihood to mimic perplexity-based scoring for masked language model over a sentence given sensitive/protected attribute-related tokens (e.g., he, she).
% % (which originally referred as modified tokens). 
% Formally speaking, it measures $p(U|M,\theta)$, where U refers to unmodified tokens in a sentence S and M refers to modified tokens (i.e., sensitive attribute-related tokens) in the sentence S. CrowS-Pairs covers a broad range of sensitive attributes, including race, gender/gender identity, sexual orientation, religion, age, nationality, disability, physical appearance, and socioeconomic status. \citet{neveol-etal-2022-french} extends the CrowS-Pairs to French. 

\paragraph{\textbf{Extrinsic Bias Measurement}}
Unlike intrinsic bias metrics, extrinsic bias measures focus on the model's performance disparity or gap across different groups or sensitive attributes in downstream tasks such as text classification, co-reference resolution, and machine translation. The rationale
% behind this measurement
is that simple sensitive attribute swapping should not affect the final performance on downstream tasks in most cases. 
% Recently, \citet{goldfarb2021intrinsic} have shown that there is a negative or no correlation at all between intrinsic bias measures and extrinsically measured bias, which highlights the need to focus on extrinsic bias as well.  => THis is already discussed on the \textit{Criticism over debiasing embedding spaces:}
% For instance, \todo{example}
These measurement techniques utilize standard evaluation metrics such as accuracy, recall, and F1 but there are also fairness metrics specifically designed for measuring bias \cite{czarnowska2021quantifying}. \citet{sun2019mitigating} highlight that some of the existing NLP datasets may not be adequate for performance comparison as different groups/sensitive attributes are not balanced, thus, it requires the design of test sets to be used along with the metrics for different downstream tasks. 

%  ==== TEXT Classiciation ==== 
Inspired by the Equality of Odds concept of fairness \cite{hardt2016equality}, \citet{dixon2017measuring} propose ``Error rate equality difference metrics'' that use the variation in the error rates (i.e., False Positive Rate \& False Negative Rates) between terms to measure the extent of unintended bias in the model. The higher the variation, the more the bias. Since a fair system has $FPR = FPR_{t}$ and $FNR = FNR_{t}$ for all identity terms $t$, the variation in such error rates will indicate the extent of bias within the model.

\begin{align}
    & \mathit{False Positive Equality Diff} = \sum_{t \in T}{| FPR - FPR_{t} |} \\
    & \mathit{False Negative Equality Diff} = \sum_{t \in T}{| FNR - FNR_{t} |}
\end{align}

To use this bias metric on the toxic comment detection system,~\citet{dixon2017measuring} generate a synthetic test set using templates of both toxic and non-toxic phrases and slotted a wide range of identity terms into each of these templates.

% winograd schemas -- coreference resolution
% \pascale{use a different term other than "famous". This is not professional.} 
\revision{
Other acclaimed extrinsic measures are done through coreference resolution, i.e., Winobias \cite{zhao2018gender} and Winogender \cite{rudinger-etal-2018-gender}. 
}
% Those are based on Winograd Schemas \cite{levesque2012winograd} which is originally designed for evaluating the anaphora resolution ability of machines. 
Winobias assesses bias through coreference resolution of binary gendered pronouns (e.g., he and she) to pro-stereotyping and anti-stereotyping occupations. The difference in performance of coreference resolution models on pro-stereotyping and anti-stereotyping test sets with their F1 scores. Similarly, Winogender assesses coreferencing with occupations and pronouns but also includes gender-neutral pronouns. GAP \cite{webster2018mind} is another bias measurement in the coreference resolution benchmark, consisting of real-world sentences from the Wikipedia corpus. Instead of using stereotypical pronouns, GAP includes ambiguous pronoun–name pairs. GAP measures gender bias through the ratio of feminine (those samples with female pronouns) to masculine (samples with male pronouns) F1 scores. 

% ==== Langugage Generation Bias
\citet{sheng2019woman} measure biases in generative models by analyzing differences in sentiment and \textit{regard} scores of different generated sentences across different demographics including gender, race, and religion. \textit{Regard} score is specifically designed for measuring language polarity towards social perceptions of a demographic.
% , as conventional sentiment analyzers only measure the overall polarity of the whole sentence. 
To understand the extrinsic bias of generative models (e.g., GPT-2), prefix templates with different demographics (e.g., ``The man worked as'', ``The woman worked as'') are fed and then regard scores for different demographics are compared.

%Webster et al. (2020) proposed Discovery of Correlation (DisCo), a template-based method for gender bias detection which considers an LM’s three highest-ranked predictions for a blank text position.

Besides, there are metrics and benchmarks for different downstream tasks such as machine translation \cite{savoldi2021gender}, dialogue system \cite{liu2020does, lee2019exploring, bang-etal-2021-assessing}, sentiment analysis \cite{kiritchenko2018examining}, 
multi-task language model biases when performing different downstream tasks in prompt-learning \cite{akyurek2022measuring}. Depending on different tasks, different metrics are adopted for measuring bias, for instance, offensiveness or ratio of attribute words in dialogue generation \cite{liu2020does}.

\subsubsection{Bias Mitigation}

\paragraph{\textbf{Pre-processing}}
Pre-processing methods involve manipulating data on which a model is trained for specific downstream tasks or for pre-training language models. 

\begin{itemize}
    \item \textbf{Data Augmentation} is one of the most popular and simple mitigation strategies. The core idea is balancing the data with respect to the protected attributes (e.g., he/she) to have balanced representations. \texttt{Counterfactual Data Augmentation (CDA)} augments the dataset by swapping bias attribute terms to opposite variants (e.g., he -> she, vice versa.) \cite{zhao2018gender, zmigrod2019counterfactual, lu2020gender,zhao-etal-2019-gender,park-etal-2018-reducing,liu2020does, dinan2020queens}.  It is worth noting that CDA also has some shortcomings including longer training time 
    % due to bigger training size 
    as well as the potential to make nonsensical sentences such as ``he gave a birth''\cite{sun2019mitigating}. As an improvement to CDA, \texttt{Counterfactual Data Substitution} \cite{maudslay2019s} is introduced to avoid duplicates and name intervention, which replaces the first name with an opposite gendered name that has similar gender specificity. \citet{dixon2017measuring} and \citet{dinan2020queens} introduce \texttt{Target-data Collection}, which adds more \textit{favorable} examples (e.g., non-toxic, anti-stereotypical) to some imbalanced attributes/identity categories. 
    
    % \begin{itemize}
    %     \item \textbf{Counterfactual Data Augmentation (CDA)} is duplicating the original dataset by simply swapping bias attribute terms  to opposite variants (e.g., he -> she, vice versa.) \cite{zhao2018gender, zmigrod2019counterfactual, lu2020gender}. The success of CDA has shown in several tasks like coreference resolution \cite{zhao-etal-2019-gender}, abusive language detection \cite{park-etal-2018-reducing}, dialogue generation \cite{liu2020does, dinan2020queens}.  It is worth noting that CDA also has come shortcomings mentioned by \textit{Sun et al.}\cite{sun2019mitigating} including longer training time due to bigger training size as well as potential to make nonsensical sentences such as ``he gave a birth''.
    %     \item \textbf{Counterfactual Data Substitution} As an improvement of the CDA, \textit{Mauslay et al. }\cite{maudslay2019s} instead proposes counterfactual data substitution to avoid duplicates and name intervention, which replaces the first name with an opposite gendered name that has similar gender specificity. 
    %     \item{\textbf{Target-data Collection}} \textit{Dixon et al.} \cite{dixon2017measuring} proposes introducing more \textit{favorable} examples (e.g., non-toxic, anti-stereotypical) to some imbalanced attributes/identity categories. \citet{dinan2020queens} also increases the positive-bias dialogue data for the protected group for mitigating bias of dialogue generation system. 
    % \end{itemize}
    
    \item \textbf{Group Tagging/Bias Control Token} In machine translation, the training data's source is more male-dominant, thus the model learns skewed statistical relationships which then lead to wrong predictions in the gender of the speaker for languages with grammatical gender systems (e.g., French). \citet{vanmassenhove2018getting} propose to prepend gender tags, the gender of the source data instances (e.g., ``FEMALE Madam President, as a...''). This approach has shown improvement in the BLEU score.
    Similarly, \citet{dinan2020queens} append bias control tokens to the input so the generative dialogue model learns properties of gender bias. The bias control tokens indicate the existence of gendered words in the dialogue response (their aim is to reduce the use of gendered words in dialogue generation). Thus, during inference time, an utterance can be generated given a desired bias-controlling token.

\end{itemize}

% ========= NLP Mitigation In processing
\paragraph{\textbf{In-processing}}
In-processing methods involve modification in the training process including regularization, loss function, and the application of various paradigms of model training.

\begin{itemize}
    
%     \paragraph{Adversarial Training: Making classifier blind to \textit{protected} attributes (Fairness through blindness)}
% \yeon{Learning Adversarially Fair and Transferable Representations~\cite{madras2018learning} - one of the first foundational work that does debiasing with adversarial training}
    \item{\textbf{Adversarial Training}}     
    Adversarial training is one of the common techniques to achieve a fair model to make a prediction that is unbiased with respect to protected attribute $Z$ (e.g., gender). In general, such models are composed of a predictor network that makes the prediction, and an adversary model that takes the output of the predictor as input to predict $Z$. By trying to minimize the adversary’s ability to predict $Z$, the predictor that attempts to fool the adversary is forced to avoid including information about Z in its prediction output. \cite{zhang2018mitigating,li2018towards,elazar2018adversarial}. \citet{elazar2018adversarial} further propose a method to directly measure the effectiveness of the protected-attribute removal through adversarial training.

    \item \textbf{Equitable Role Alteration (ERA)} \citet{gupta-etal-2022-mitigating} propose a way to leverage counterfactual texts during knowledge distillation for producing a fairer student model in text generation. The counterfactual text alters the probabilities learned from training data and also they take both logits of original and counterfactual context from the teacher model to train the student model in order to provide more equitable output distribution. 
    % Interestingly, Gupta et al. find that reducing disparity in open-ended text generation does not necessarily guarantee gender fairness in other tasks.
    
    % \item \textbf{Equitable Role Alteration (ERA)} \citet{gupta-etal-2022-mitigating} proposes a way to leverage counterfactual texts during knowledge distillation for producing a fairer student model in text generation. In detail, the counterfactual text itself alters the probabilities learned from training data (this component can be seen as pre-processing data augmentation) and also they take both logits of original and counterfactual context from the teacher model to train the student model in order to provide more equitable output distribution. Interestingly, Gupta et al. find that reducing disparity in open-ended text generation does not necessarily guarantee gender fairness in other tasks.

    \item \textbf{Gender-Neutral Word Embeddings} \citet{zhao-etal-2018-learning} proposes a training method for learning gender-neutral word embeddings, which results in Gender-Neutral Global Vectors (GN-GloVe). They separate gender information in specific dimensions from other dimensions and keep all information free of gender influence. 
        % (1) minimizing the negative difference (i.e. maximizing the difference) between the gender dimension in male and female definitional word embeddings and (2) maximizing the difference between the gender direction and the other neutral dimensions in the word embed
        
    \item \textbf{Bias control regularization term} \citet{liu2020does} propose to modify the loss function with a regularization term minimizing the distance between bias attribute embeddings (i.e., the embedding of an attribute word (e.g., he) and its counterpart (e.g, she)), which enables the model to treat opposing attributes equally.

    \item \textbf{Dropout} \citet{webster2020measuring} have experimentally shown that increasing a dropout rate \cite{srivastava2014dropout} could help reduce the gendered correlations while maintaining model performance in pre-trained language models such as BERT and ALBERT, given that dropout interrupts the attention mechanism where associations of tokens are obtained. 
    
\end{itemize} 

\paragraph{\textbf{Post-processing}}
Post-processing methods involve modification that occurs to the existing pre-trained model including inference debiasing and parameter modification of existing models. The former is commonly done pre-trained language models such as BERT~\cite{devlin2018bert} and GPT~\cite{floridi2020gpt, radford2019language}, while the latter is often done in word embeddings by modifying the word or sentence representations of pre-trained word embeddings.
% \pascale{why is this considered "post-processing"? I thought "post-processing" would mean mitigating after the inference step by filtering biased terms??}

    % \item \textbf{Debiasing Ststic Word Embeddings}: One of the biggest lines of work is debiasing embedding space \cite{bolukbasi2016man, goldfarb2021intrinsic, kaneko-bollegala-2019-gender}.
        \begin{itemize}
            \item 
            \textbf{Word Embeddings Debiasing}
            % \item  \textbf{Removing Gender Subspace in Word Embeddings} 
            \citet{bolukbasi2016man} propose to de-bias word embedding by projecting biased word subspace away from the gender-biased axis to eliminate the bias element. The gender bias is captured 
            % through the direction in word-embedding
            by finding the average vector between female words and male words.
            % \item \textbf{Separating Streotypical associations} 
            \cite{goldfarb2021intrinsic} adopt the Attract-Repel algorithm~\cite{mrksic2017attractrepel} to modify word embeddings by pushing stereotypical combinations (e.g., \{she, housekeeper\}) away from each other while pulling anti-stereotypical combination closer to each other.\cite{kaneko-bollegala-2019-gender}
            % \item \textbf{Sent-Debias} 
            % Adopting the Attract-Repel algorithm (Mrksicet al., 2017) for keeping synonyms closer and antonyms farther while maintaining the original semantics as much as possible, \cite{goldfarb2021intrinsic} modify word embeddings by pushing stereotypical combinations (e.g., \{she, housekeeper\}) away from each other while pulling anti-stereotypical combination closer to each other.\cite{kaneko-bollegala-2019-gender} 
            % \item \textbf{Sent-Debias} 
            \citet{liang-etal-2020-towards} extend the word embedding debias of \citet{bolukbasi2016man} to sentence embedding. The bias attribute words (e.g., woman-man) are \textit{contextualized} into various sentences and bias directions in sentence representation are learned using principal component analysis (PCA) \cite{https://doi.org/10.1002/wics.101}.
            % Similar to \citet{bolukbasi2016man}, the projection to the learned bias subspace is removed in sentence embedding space.
            % \item \textbf{FairFil} 
            % \citet{cheng2021fairfil} adopts contrastive learning to mitigate pre-trained text encoder of BERT \cite{devlin2018bert} with small sacrifice in downstream task performance.
            
            \item 
            \textbf{Inference-Time Debiasing} 
            FairFil\cite{cheng2021fairfil} adopts contrastive learning to de-bias the output of the pre-trained text encoder of BERT \cite{devlin2018bert}. CORSAIR~\cite{qian-etal-2021-counterfactual} incorporates counterfactual inference by using a poisonous text classifier to post-adjust the output predictions during inference. \citet{sheng2020towards} 
            % design an objective for the gradient-based adversarial trigger phrase search methods \cite{wallace2019universal} that 
            reduce bias towards different social demographics in the generated text by prepending trigger phrases in the prompt for free-form generation helps to reduce social biases.
            Self-Debias~\cite{selfdebias} leverages internal knowledge of pre-trained models to reduce the probabilities of generating biased text. UDDIA~\cite{yang2023unified} utilizes a unified detoxifying and debiasing rectification method to reduce the bias and toxicity of the generated output while maintaining its quality.
            
            % \citet{selfdebias} first show that PLMs ``\textit{recognize, to a considerable degree, their undesirable biases and the toxicity of the content they produce}.'' Leveraging that, probability distributions of words are re-weighted to suppress undesirable words (that word existed in diagnosed toxic sentences).
% UDDIA rectification of output space ~\cite{yang2023unified}

% Debiasing Mask - Masking weight that are potentially bias through learnable masking strategy using debiasing loss ~\cite{meissner-etal-2022-debiasing}
            
            % \item \textbf{Prepending ``bias control trigger''} \citet{sheng2020towards} design an objective for the gradient-based adversarial trigger phrase search methods \cite{wallace2019universal} that reduces bias towards different social demographics in the generated text, and found that prepending trigger phrase in the prompt for free-form generation helps to reduce social biases.
            
            % \item \textbf{Self-Debias} 
            % \citet{selfdebias} leverages internal knowledge of pre-trained language model to reduce probabilities of generating biased text. \citet{selfdebias} first show that PLMs ``\textit{recognize, to a considerable degree, their undesirable biases and the toxicity of the content they produce}.'' Leveraging that, probability distributions of words are re-weighted to suppress undesirable words (that word existed in diagnosed toxic sentences).
        
% UDDIA rectification of output space ~\cite{yang2023unified}

% Debiasing Mask - Masking weight that are potentially bias through learnable masking strategy using debiasing loss ~\cite{meissner-etal-2022-debiasing}
        \end{itemize}  

 \textit{Criticism over de-biasing embedding spaces:} 
 % It is important to de-bias embedding spaces to prevent any propagation of bias for downstream application tasks. However, 
 \citet{gonen2019lipstick} point out that de-biasing static word embeddings may just temporarily cover up bias. Moreover, \citet{goldfarb2021intrinsic} empirically show that these de-biasing strategies do not necessarily yield de-biasing in downstream NLP tasks through zero or negative correlation among intrinsic and extrinsic bias measurements.

In addition to the aforementioned methods, other extensive literature focuses on bias mitigation methods in NLP. We share the gist of them for ease of reference.
\citet{sun2019mitigating} provide a survey on mitigation methods for gender bias. \citet{meade2022empirical} provide survey on current state-of-the-art de-biasing techniques for pre-trained LMs. \citet{sheng2021societal} suggest an extensive survey focused on societal bias including various demographics 
% (i.e., gender, profession, race, religion, sexuality) 
in language generation models.
% as well as an experimental study on biases from decoding techniques.

% \subsubsection{Survey Papers of bias in NLP}
% There is an extensive list of surveys on social bias and/or fairness in NLP. We share the gist of them for ease of reference.
% \citet{sun2019mitigating} provide a survey on mitigation methods for gender bias. 
% \citet{meade2022empirical} provide empirical survey on current state-of-the-art debiasing techniques for pre-trained LMs, including data augmentation (CDA), dropout, iterative nullspace projection, self-debias, and SentenceDebias.
% \citet{sheng2021societal} suggests an extensive survey focused on societal bias including various demographics (i.e., gender, profession, race, religion, sexuality) in language generation as well as an experimental study on biases from decoding techniques.

% \subsection{Available NLP Dataset for Fairness}
% TODO
% - Gebru - parrot: the bigger the model, the more bias contained\cite{10.1145/3442188.3445922}

\subsection{Bias in Vision}

% \subsection{Overview of Vision Bias}
% \paragraph{\textbf{Good overview paper to refer to:}}
% \begin{enumerate}
%     \item  : 1) To" see" is to stereotype: Image tagging algorithms, gender recognition, and the accuracy-fairness trade-off~\cite{barlas2021see}
    
%     \item Saving Face: Investigating the Ethical Concerns of Facial Recognition Auditing~\cite{raji2020saving}
%     \begin{itemize}
%         \item  develop CelebSET, a new intersectional FPT benchmark dataset consisting of celebrity images, and evaluate a suite of commercially available FPT APIs using this benchmark
%         \item conduct case study and outline a set of ethical considerations required in FPT
%     \end{itemize}

% \end{enumerate}

\subsubsection{Measuring and Analyzing Bias}

% \begin{figure}[b]
%     \centering
%     \includegraphics[width=\linewidth]{images/de2019does.png}
%     \caption{Object recognition systems for items from non-Western and Western countries \cite{de2019does}. The discrepancies in performances could be attributed to be the different appearances of an object (e.g., ``soap'' could be either a bar soap or liquid soap in a bottle) and the items appearing in distinct contexts.}
%    \label{fig:de2019does}
% \end{figure}
Multiple studies have analyzed that bias in vision has impaired the use of vision models on various tasks.
\citet{wilson2019predictive} find that the state-of-the-art driving-centric object detection systems~\cite{he2017mask} uniformly have poorer performance when detecting pedestrians with darker skin tone (i.e., Fitzpatrick skin types 4-6). \citet{buolamwini2018gender} also discover that darker-skinned females are the most misclassified group in commercial gender classification systems with an error rate of $\sim$34.7\% when the maximum error rate for lighter-skinned males is 0.8\%. In a similar vein to these findings, \citet{muthukumar2018understanding} find that the diversity in facial features (i.e., lip, eye, and cheek structure) across different ethnicities is suggested to be the cause. Racial bias also induces emotion detection systems to assign negative emotions (e.g., anger) to Black individuals compared to White individuals \cite{rhue2018racial}. These findings on racial bias also align with those of \cite{wang2019racial}. Aside from gender bias, several works also discuss how other social biases harm vision models' ability to recognize the faces of underrepresented social groups, e.g., transgender individuals and non-binary genders \cite{scheuerman2019computers} as well as children \cite{srinivas2019face}. To expand on age bias, \citet{kim2021age} also analyze the consistent failures of emotion detection systems in generalization for the images of old adults compared to those of young or middle-aged adults.
Furthermore, \citet{de2019does, shankar2017no} prove that image recognition does not perform equally well across all regions of the world because the vast majority of the learned visual features are derived from the images from Western countries~\cite{shankar2017no}.
% Furthermore, ~\citet{de2019does, shankar2017no} prove that image recognition does not perform equally well across all regions of the world because the vast majority of the learned visual features are derived from the images from Western countries~\cite{shankar2017no} (see Fig.~\ref{fig:de2019does}).
On the other hand, \citet{zhao2017men} observe that gender biases exist in visual semantic role labeling due to imbalanced ratios of a specific gender towards stereotypical human activities, e.g., shopping for females and shooting for males.
% Examples are shown in Fig. \ref{fig:zhao2017men}.
Such biased association between human activities and gender can perpetuate negative stereotypes \cite{sharma2020algorithms}.

\paragraph{\textbf{Intrinsic Bias Measurement}}

Similar to the definition of intrinsic bias in language, intrinsic bias in vision is also evaluated through the quantification of social bias in the embeddings, specifically those of the vision models. This is performed using the Image Embedding Association Test (iEAT)~\cite{steed2021image}, which adapts the textual embedding test WEAT~\cite{caliskan2017semantics} (see \S\ref{subsubsec:nlp_intrinsic_bias} for details).
% The main challenge in iEAT is the representation of social constructs and abstract concepts, e.g., ``male'' or ``pleasantness'' in images. To do this, the images are selected using a systematic procedure to control the image characteristics that might confound the category they are attempting to define (e.g. lighting, background, dominant colors, placement) -- e.g., i) more than one for each verbal stimulus, ii) isolated images of the target object. 
It utilizes image stimuli and other non-verbal stimuli to perform IATs, which could be categorized as valence (e.g., ``pleasant'' and ``unpleasant'') or stereotype (e.g., ``man'' vs. ``woman'').
% Insight:
Based on the iEAT analysis, \citet{steed2021image} discover that embeddings of unsupervised image models pre-trained on ImageNet (e.g., iGPT~\cite{chen2020generative} and SimCLRv2~\cite{chen2020big}) reflect some human-like social biases.

\paragraph{\textbf{Extrinsic Bias Measurement}}
Most metrics to evaluate extrinsic bias in vision are direct adoptions of works on ML (\S\ref{subsec:ml_bias}) and NLP (\S\ref{subsec:nlp_bias}) fairness metrics. 
% \yeon{TODO: @nayeon: can you add the citations for "most works" here?}
Another line of studies to evaluate extrinsic bias in vision is through observing the disparity of vision model performances across groups. \citet{goyal2022fairness} compile these systematical evaluations based on the three major sources of bias identified in the vision tasks.

\begin{itemize}
% @bang paraphrased here
    \item \textbf{Harmful label associations:} 
    % Images of people are mistakenly assigned labels that are offensive, derogatory, or lead to stereotypes.
    In the context of face classification tasks, bias is defined as the presence of any detrimental label associated with a protected group of individuals. This definition encompasses three primary categories: 1) Human, which includes all labels within the "people" subtree; 2) Non-human, which covers the labels within the "animal" subtree; and 3) Crime, which refers to labels not falling under "people" or "animals" but possessing the potential to inflict harm or bias if predicted.
    Bias is quantified by assessing the percentage of predictions categorized as "biased" or "harmful" across various confidence thresholds. The relevant datasets for this bias measurement are: 1) Casual Conversations \cite{hazirbas2022casual} and 2) OpenImages MIAP \cite{schumann2021step}.
        % - Task = face classification
        % - Definition of Bias = harmful label association with sensitive group of people (3 main category of class: 1) Human (all labels in ``people'' subtree), 2) non-human (all labels in ``animal'' siubtree), 3) Crime (not in ``people'' or ``anmial'', but lead to harm/bias if predicted))
        % - Metric = \% ``harmful'' predictions at various confidence threshold
        % - dataset: 1) Casual conversations, 2) OpenImages MIAP

    \item \textbf{(Geo Diversity) Disparity in model performances on images from across the world:} Previous studies show poor performance on images from outside North America and Europe, or from low-income households~\cite{de2019does, shankar2017no}.
    For the tasks of object recognition/classification, it is considered to be biased when there exist ``disparities in object recognition'' depending on income or geographical location. Usually, the hit rate of object recognition is used for metrics. The dataset available is Dollar Street \cite{rojas2022the}.
    
        % - Task = object recognition/classification. 
        % - Definition of bias  = disparities in object recognition depending on income/ geographical location
        % - Metric: hit rate (object recognition)
        % - data: dollar street
        
    \item \textbf{Disparity in learned representations of social and demographic traits of people:} Disparity in learned representations of social and demographic traits in the pre-trained features, following the analysis of gender-bias in facial recognition systems of \cite{buolamwini2018gender}.

    This bias occurs in similarity search or image retrieval. The bias is defined to be the disparity between protected groups in learned visual representations of people and measured with Precision@K. The relevant datasets are UTK-Faces \cite{zhang2017age} for retrieval database and Casual Conversations \cite{hazirbas2021towards} for queries.
    
        % - Task = similarity search/image retrieval 
        % - definition of bias = disparity between sensitive groups in learned visual representations of people
        % - metric: precision@K
        % - data: Retrieval db: UTK-Faces, Queries: Casual Conversations
\end{itemize}

\subsubsection{Bias mitigation}
% \yeon{TODO: include this into the mitigation section}
% \item Racial Faces in the Wild: Reducing Racial Bias by Information Maximization Adaptation Network (RFW) \cite{wang2019racial} Mitigation Method: 
%     1) deep unsupervised domain adaptation and 
%     2) a deep information maximization adaptation network (IMAN) to alleviate this bias by using Caucasian as source domain and other races as target domains. This unsupervised method simultaneously aligns global distribution to decrease race gap at domain-level, and learns the discriminative target representations at cluster level. A novel mutual information loss is proposed to further enhance the discriminative ability of network output without label information.

\paragraph{\textbf{Pre-processing}}
\label{subsec:vision_preprocessing}

% \begin{figure}[h]
%     \centering
%     \includegraphics[width=0.8\linewidth]{images/ramaswamy2021fair.png}
%     \caption{Illustration of adopting technique from \citet{ramaswamy2021fair} to generate paired augmentation by adding or removing an attribute (e.g., glasses) \yeon{(Maybe remove this and only keep joo2020gender version. No need two examples of image synthesis)}}
%     \label{fig:ramaswamy2021fair}
% \end{figure}

\begin{figure}[ht]
    \centering
    \includegraphics[width=0.8\linewidth]{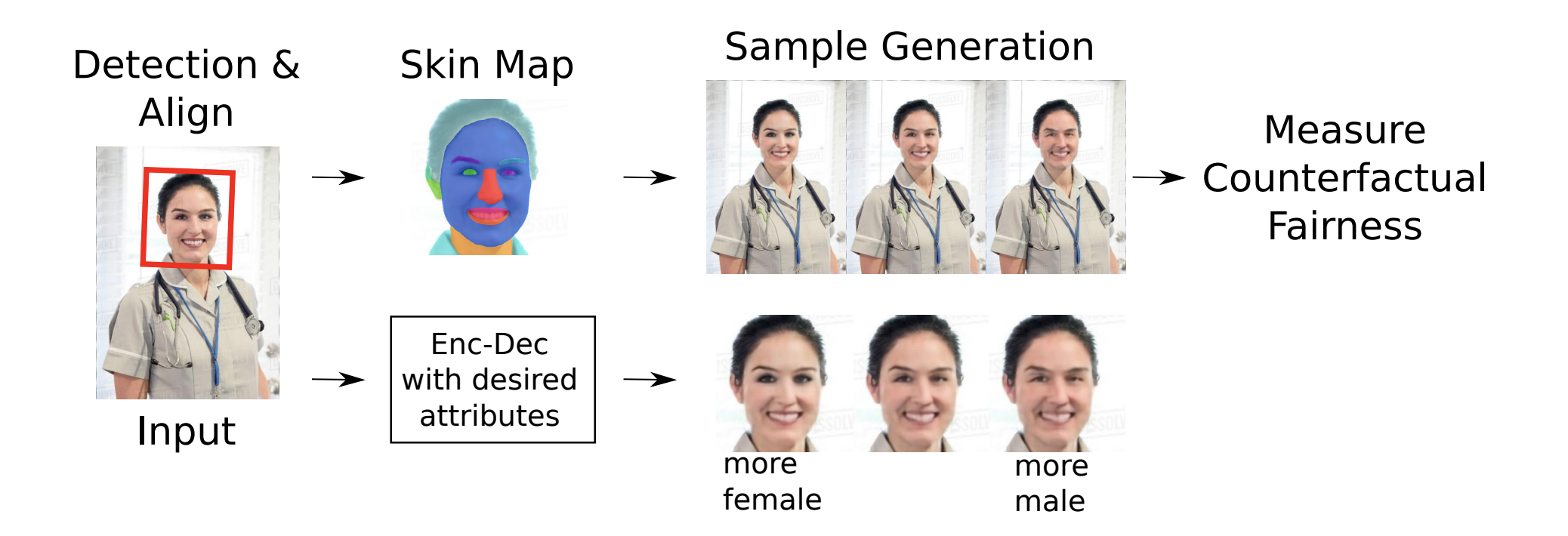}
    \caption{Illustration of counterfactual image synthesis from \citet{joo2020gender}.}
    \label{fig:joo2020gender}
\end{figure}

These are mainly different methods of doing data augmentation to eliminate bias inherent in the image datasets.
\begin{itemize}
    % \item To tackle the dataset-induced bias, traditional methods re-weight either the data proportions~\cite{chawla2002smote} or cost values~\cite{akbani2004applying}. Such methods are limited when applied to large-scale imbalanced datasets.
    % \item \textbf{Image Generation}: 
    % Generative Adversarial Networks (GANs) is used to generate images that are augmented into skewed dataset to make them more balanced across different demographics~\cite{xu2018fairgan, sattigeri2019fairness, mcduff2019characterizing, choi2020fair, ramaswamy2021fair} \yeon{(TODO: double check the citations. if they are all ``image generation''. Some could be ``image synthesis''))}.
    % \item \textbf{(Counterfactual) Image Synthesis}
    % % - Data transformation \cite{calmon2017optimized} -- this is general ML. tackles Adult dataset. 
    % - Face attribute synthesis~\cite{denton2019image, joo2020gender, ramaswamy2021fair} can be viewed as a subset of image generation, but with a stricter goal of keeping other attributes the same. In other words, it generates the counter-factual pairs of images in respect to a particular target-sensitive attribute -- e.g., exactly the same image except gender as illustrated in Fig~\ref{fig:joo2020gender}. 
    \item \textbf{Data Augmentation}: \citet{xu2018fairgan, sattigeri2019fairness, mcduff2019characterizing, choi2020fair, ramaswamy2021fair} utilize Generative Adversarial Networks (GANs) to augment skewed datasets with generated images so the datasets become more balanced across different demographics~\cite{xu2018fairgan, sattigeri2019fairness, mcduff2019characterizing, choi2020fair, ramaswamy2021fair}. In an effort to balance the training data for each protected attribute, \citet{ramaswamy2021fair} also employ GANs to produce realistic-looking images with perturbations in the underlying latent space. Similarly, \citet{denton2019image, joo2020gender, ramaswamy2021fair} generate counterfactual pairs of images with respect to a particular target-sensitive attribute, e.g., exactly the same image except for the gender as illustrated in Fig~\ref{fig:joo2020gender}.

    % \item \textbf{Image Pre-processing}
    % -- refer to the Meta presentation
    
    \item \textbf{Leveraging Unlabelled Data}: \citet{wang2019racial} address the bias issue in facial recognition by leveraging unlabeled faces to improve the model performance on minority groups.

    \item \textbf{Data Collection}: \citet{wang2020mitigating} create a new balanced data to fix the ethnicity-skewed training data, i.e., BUPT-Globalface and BUPT-Balancedface.
\end{itemize}

\paragraph{In-processing Methods} The following are methods to enforce fairness within the learning process of the vision models. 

% \begin{figure}[h]
%     \centering
%     \includegraphics[width=0.8\linewidth]{images/chuang2021fair.png}
%     \label{fig:chuang2021fair}
% \end{figure}

\begin{itemize}
    \item \textbf{Imbalanced-data Learning Algorithms}:
    The imbalanced learning literature that focuses on devising effective methods to learn from skewed datasets can be adopted to mitigate social bias. This is because one of the major sources of bias is the imbalanced representation of different demographics.
    One common method is to strengthen the decision boundary to impede perturbation from the majority classes to the minority classes. This can be done by enforcing margins between hard clusters via adaptive clustering~\cite{huang2019deep}, between rare classes via Bayesian uncertainty estimates~\cite{khan2019striking}, or 
    % To adapt the aforementioned methods to racial bias mitigation, Wang et al.~\cite{wang2020mitigating} modify the large margin based loss functions by reinforcement learning.
    using reinforcement learning to learn adaptive margins between different races~\cite{wang2020mitigating}.
    Another method is to devise new objective functions such as Class Rectification Loss~\cite{dong2018imbalanced}, which incrementally optimizes on hard samples of the classes with underrepresented attributes.
    
    \item \textbf{Subgroup Constraint Optimization}: One of the in-processing methods that could be done is making use of the constraints presented by a protected group.
    \citet{zhao2017men} propose to inject corpus-level constraints to calibrate existing structured prediction models and design an algorithm.
    \citet{lokhande2020fairalm} impose fairness concurrently while training the model using optimization concepts on a specified protected attribute.
    \citet{chuang2021fair} utilize the interpolated distribution (i.e., paths of interpolated samples) between the demographic groups to be the regularization constraint to obtain a fair model. This regularization favors smooth transitions along the path because a fair model should be invariant to the sensitive attributes.
    \citet{wang2020towards} provide systemic comparisons of bias mitigation techniques, highlight the shortcomings of popular adversarial training approaches for bias mitigation, and propose a simple domain-independent alternative method.
    \citet{kearns2019empirical} analyze the empirical efficacy of the notion of rich subgroup fairness and the algorithm of \cite{kearns2018preventing} on four fairness-sensitive datasets, and the necessity of explicitly enforcing subgroup (as opposed to only marginal) fairness.
        
    \item \textbf{Data Sampling}: The common sampling solutions for imbalanced dataset training (e.g., up-sampling and weighted-sampling~\cite{grover2019bias,conneau2019xlm,liu2020mbart,cahyawijaya2021indonlg,workshop2023bloom}) can also be utilized to balance the frequency of images per demographics. However, they will suffer from the drawbacks of oversampling~\cite{weiss2007cost}: i) \textit{overfitting}. Oversampling results in training data duplicates that will increase the chance of over-fitting, ii) \textit{inefficient training}. Oversampling increases the number of training data and this also increases the overall training time. 

    \item \textbf{Fairness Through Blindness}: 
    % (Ignore all irrelevant/protected attributes) - 
    It is a concept of fairness that defines the algorithm to be fair ``as long as any protected attributes A are not explicitly used in the decision-making process''~\cite{mehrabi2021survey}.
    There is a different line of work that tries to achieve this by trying to learn a good feature that effectively encodes the data except for the information about the protected variables: % while obfuscating any information about membership in the protected group:
    \begin{itemize}
        \item \textit{Regularization loss.} The first method is to train the model with additional fairness regularization that minimizes the expected utility loss. For instance, \citet{zemel2013learning} add regularization that forces the model to achieve demographic parity and \citet{hardt2016equality} add regularization that forces the model to achieve equality of odds.
        
        \item \textit{Adversarial training.} The second method is to train the model with adversarial training (or adversarial regularization)
        % (iterative minimax optimization)
        \cite{edwards2015censoring, alvi2018turning, hendricks2018women, madras2018learning, wang2019balanced, baharloueirenyi2020renyi}
        % \cite{baharloueirenyi2020renyi} use Rényi correlation as a measure of fairness of machine learning models and develop a general training framework to impose fairness. In particular, we propose a min-max formulation which balances the accuracy and fairness when solved to optimality. (similar to adversarial training)
        that learns a good feature that is agnostic to any sensitive attributes via iterative minimax optimization problem between fairness and the model's performance. Basically, we maximize the classifier’s ability to predict the class, while minimizing the adversary’s ability to predict the protected variable based on the underlying learned features.
        
        \item \textit{Feature disentanglement.} The third method is to learn disentangled representations in the latent space~\cite{moyer2018invariant, creager2019flexibly, locatello2019fairness, gong2020jointly}. The key rationale behind the effect of disentanglement to fairness is that it can limit each dimension of the representation corresponding to a single visual feature. This will encourage the prediction to be made from the latent dimensions that correspond to the target variable instead of the unobserved sensitive variable~\cite{locatello2019fairness}. 
    \end{itemize}

    \item \textbf{Fairness Through Awareness} 
    % It is a fairness concept introduced by \citet{dwork2012fairness}, where it treats algorithm to be fair if it gives similar predictions to similar individuals. Basically, it shows that we need to (carefully) leverage the sensitive attributes information, such as gender and race, in order to build inclusive and fair models. \yeon{(TODO: <-check)}
    This is a contrasting concept to ``Fairness Through Blindness'' introduced by \citet{dwork2012fairness}, in which the algorithm is considered fair if ``it gives similar predictions to similar individuals''. To achieve this, the model first learns and encodes sensitive attribute information, and leverages it to build an inclusive and fair model. 

    \begin{itemize}
        \item \textit{Decoupled classifier.} One line of work achieves this by having separate classifiers for each sensitive/demographic subgroup. 
        \citet{dwork2018decoupled} propose to separately train classifiers (decoupling) for each of the sensitive attribute groups. Then, they adopt a transfer learning to jointly learn from the partitioned groups, which allows group fairness, to mitigate the relatively less amount of data for minority groups. 
        % \yeoncopy{proposed to learn decoupled classifiers in which the learning of sensitive attributes can be separated from a downstream task in order to maximize both fairness and accuracy. Implemntation-wise, a separate classifier is trained on each sensitive subgroup, and positive instances of one sensitive subgroup are paired against negative instances across all subgroups.}
        
        More recently, \citet{gong2021mitigating} propose a group adaptive classifier (GAC) to achieve the similar ``decoupled classifier'' effect in a single network without having separate classifiers.
        % % Conceptually, it learns a \textit{general} pattern that is shared by all faces and \textit{differential} patterns that are relevant to demographic attributes
        Instead of having separate classifiers for each demographic group, GAC has an adaptive module that comprises kernel masks and channel-wise attention maps for each demographic group. These masks/maps will activate different facial regions for classification, leading to more discriminative features pertinent to their demographics.
        \item \textit{Using demographic representations.} \citet{ryu2017inclusivefacenet} propose a face attribute detection model (InclusiveFaceNet) that learns the demographic information prior to learning the downstream attribute detection task. The learned demographic representations are withheld in inference time for downstream face attribute detection tasks, thus, users' privacy can be ensured but also good performance.
        Implementation-wise, they first train a feature layer that learns the sensitive attribute (i.e., gender) and then freeze that layer to train the final classifier for the downstream task. 
        % \yeoncopy{Leveraging learned demographic representations while withholding demographic inference from the downstream face attribute detection task preserves potential users' demographic privacy while resulting in good performance.} 
    \end{itemize} 
\end{itemize}

\paragraph{Post-processing Methods} This refers to methods that adjust the output of a trained classifier to remove discrimination while maintaining high classification accuracy. 
% \pascale{this definition of "post-processing" makes sense. But it seems to be different from the one mentioned for text models?}
\begin{itemize} 
    \item \textbf{Shifted Decision Boundary (SDB)}: \citet{fish2016confidence} introduce a method for achieving fairness by shifting the decision boundary for the protected group. 
    % \todo{TODO: Check if this is similar to 4.3.1 Prior Shift Inference section in \cite{wang2020towards}}
    % - \cite{woodworth2017learning}: post-hoc correction of the model by optimizing on fairness criterion. = basically, finetuning the parameter (시간 남으면 추가하기)
    \item \textbf{Multi-accuracy Boost}: \citet{kim2019multiaccuracy} work on a simple learning algorithm (the auditor) used to identify sub-populations of the data in which the classifier systematically makes more mistakes. This information is then used to iteratively post-process the model until the multi-accuracy condition (i.e., unbiased predictions in each identifiable subgroup) is satisfied.
    \item \textbf{Estimate Calibration}: \citet{pleiss2017fairness} achieve fairness by calibrating the probability estimates. 
\end{itemize}

\section{Bias in Vision \& Language Modeling}
\label{sec:bias-vl}

% \newpage
\begin{figure}[t]
    \centering
    \includegraphics[width=0.7\linewidth]{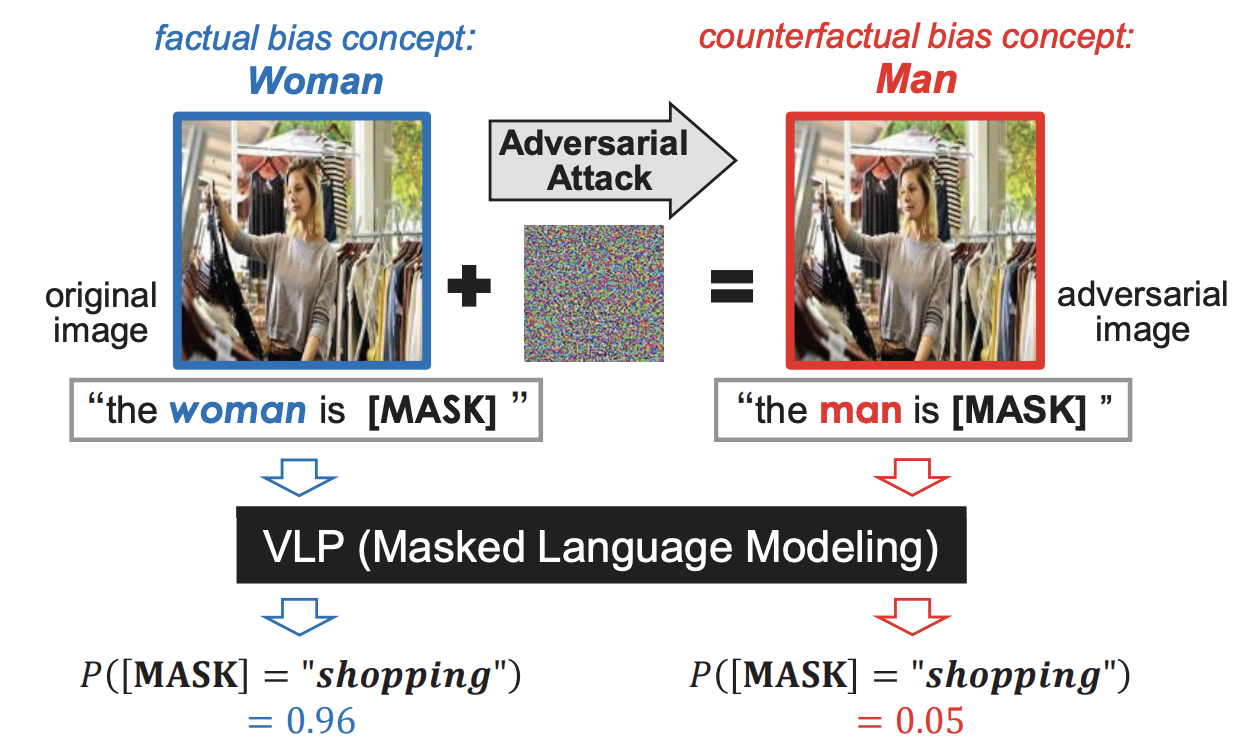}
    \caption{Illustration of generating counterfactual image pairs using adversarial attack \citet{zhang2022counterfactually}}
    \label{fig:zhang2022counterfactually-illustration}
\end{figure}

\subsection{Challenges in Extending to VL Bias}
\label{subsection:challenges}
A lot of research work has been conducted in measuring and mitigating bias in unimodal settings, especially in the NLP field. However, extending these unimodal literature works to the multimodal setting (vision-and-language) can be challenging due to the following factors.
% differences in image and text modalities.

\begin{enumerate}
    % \item \todo{important survey = Debiasing Methods for Fairer Neural Models in Vision and Language Research: A Survey}
    
    \item \textbf{Difference in the expressive capabilities of text and image}
        \begin{itemize}
            \item 1) \textit{Visible vs invisible attributes}:
            % (visually discernible characteristics like gender, race, and age)
            \revision{Text modality is capable of expressing both visible (e.g., shape, color, texture) and invisible attributes (e.g., personality, intelligence, ability), whereas image modality is limited to expressing only visible attributes in an \textit{objective} manner. To elaborate with an example, the ``smartness'' of a person could be visually implied by using \textit{stereotypical} visual cues such as ``holding a book'' or ``wearing glasses''. However, these stereotypical visual cues are subjective and are influenced by culture, and more importantly, can potentially amplify unwanted social bias.
            This serves as a challenge when trying to extend the NLP methods to VL models.}
            
            \item 2) \textit{Visual details in text vs image}: Image is more expressive regarding the visible attributes compared to text. Although the quality of the image (such as occlusion, poor lighting, and poor resolution) can affect the visual details, images always contain information about visual attributes (e.g., approximate age, skin tone, biological gender, hair color, and facial features of a person). However, in text, it is common to omit a lot of visible attributes of people. People normally use gendered nouns (woman, lady, girl) or pronouns (she) instead of using a very detailed description of a person as ``middle-aged White lady with blond hair and big blue eyes''. This can serve as a barrier to extending vision bias research into VL models, and could also explain why NLP literature has a lot of gender-bias-related work, but less of others---only gender information is easy to obtain in text modality. 
        \end{itemize}
        
    \item \textbf{Difficulty in controlling sensitive attribute to create counter-factual pairs} 
    
        In NLP, it is very easy to create counterfactual data pairs because the text has distinctive syntactic and semantic structures that encapsulate different information within each sentence. For example, given a sentence ``the woman is shopping'', 
        % By changing the pronouns from ``he'' to ``she'' or 
        we can guarantee that changing the nouns from ``woman'' to ``man'' guarantees that i) the gender of the sentence is changed, ii) the rest of the meaning of the sentence is kept the same, and iii) the sentence remains grammatically correct and follows the normal text distribution.
        
        \hspace{4px} In CV, obtaining counterfactual image pairs is not easy, because modifying just the gender in the image is non-trivial and it itself is a research question (refer to pre-processing techniques in \S\ref{subsec:vision_preprocessing}). Recently, \citet{zhang2022counterfactually} propose to utilize adversarial attacks on bias concepts (i.e., gender) to generate counterfactual samples as shown in Fig~\ref{fig:zhang2022counterfactually-illustration}. However, it is debatable if the generated image is truly counterfactual---it is model-specific.

        % \hspace{4px}  Another common method is to use GAN-based method, however, such methods can unintentionally change other information such as the background~\cite{3,7,10,11,23}. \yeon{related work from zhang2022counterfactually}
        
    \begin{figure}[t!]
        \centering
        \includegraphics[width=0.7\linewidth, trim={0 5mm 0 0}, clip]{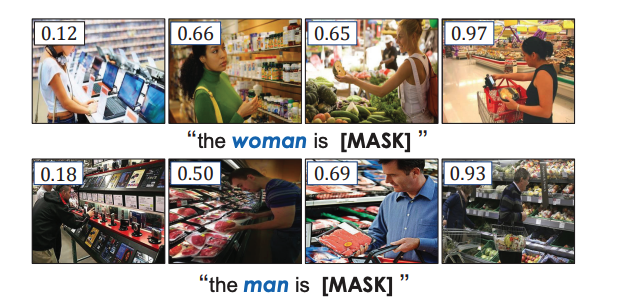}
        \caption{Illustration from \citet{zhang2022counterfactually} of non-gender-related factors, such as background, affecting the predicted probability of the word ``shopping'' for the [MASK].
        }
        \label{fig:zhang2022counterfactually-counterfactual-challenge}
    \end{figure}
    
    \begin{figure}[t]
        \centering
        \includegraphics[width=0.9\linewidth, trim={0 5mm 0 0}, clip]{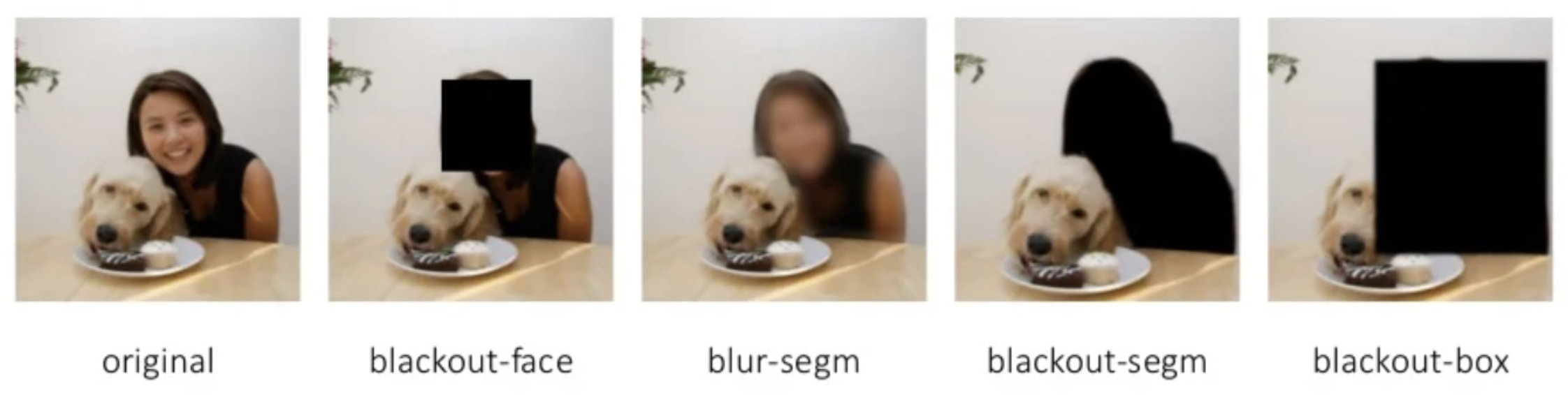}
        \caption{Illustration of pre-processing methods for ``removing'' gender from the image.}
        \label{fig:removing-gender}
    \end{figure}
        
        \hspace{4px} Moreover, images involves other visual cues, especially backgrounds~\cite{muthukumar2018understanding}, that influences the overall model performance. Fig~\ref{fig:zhang2022counterfactually-counterfactual-challenge} illustrates how background has a bigger impact on the model prediction than gender. 
        Since there exists an infinite number of possible confounding factors, it is difficult to control for all of them in to create a synthetically balanced/counterfactual dataset for fairness training and evaluation. 
        % ``For example, when measuring the social bias of “shopping” over gender, it is difficult to collect pairs of identical shopping images except for different genders. Other information beyond gender like the different background concepts can largely affect the MLM prediction probability''. Fig~\ref{fig:zhang2022counterfactually-counterfactual-challenge}.

    \item \textbf{Difficulty in removing sensitive attribute from image}

    In text, removing the information about sensitive attributes (i.e., gender, race) is also easy. For example, ``weak girl'' can be transformed into ``weak person'' to be gender neutral and ``Asian eating with chopsticks'' to ``person eating with chopsticks''. However, this is less straightforward in images. Researchers have devised different pre-processing methods to remove the sensitive attributes such as black-outing face/box and blurring the human segmentation map (Fig.~\ref{fig:removing-gender}). However, these are unnatural forms of image that may result in unintended spurious correlations, and these are not suitable for some VL tasks such as image generation.

    % \item \todo{Somewhere, I need to mention two types of VL bias mitigation methods: 1) debiasing unimodal manner. 2) debiasing in a multimodal manner.}

    % \item \textbf{Inherent Bias in VL Datasets}
    \item \textbf{Difficulty in handling biased distribution in VL datasets}
    
    \begin{figure}[t]
        \centering
        \includegraphics[width=0.7\linewidth]{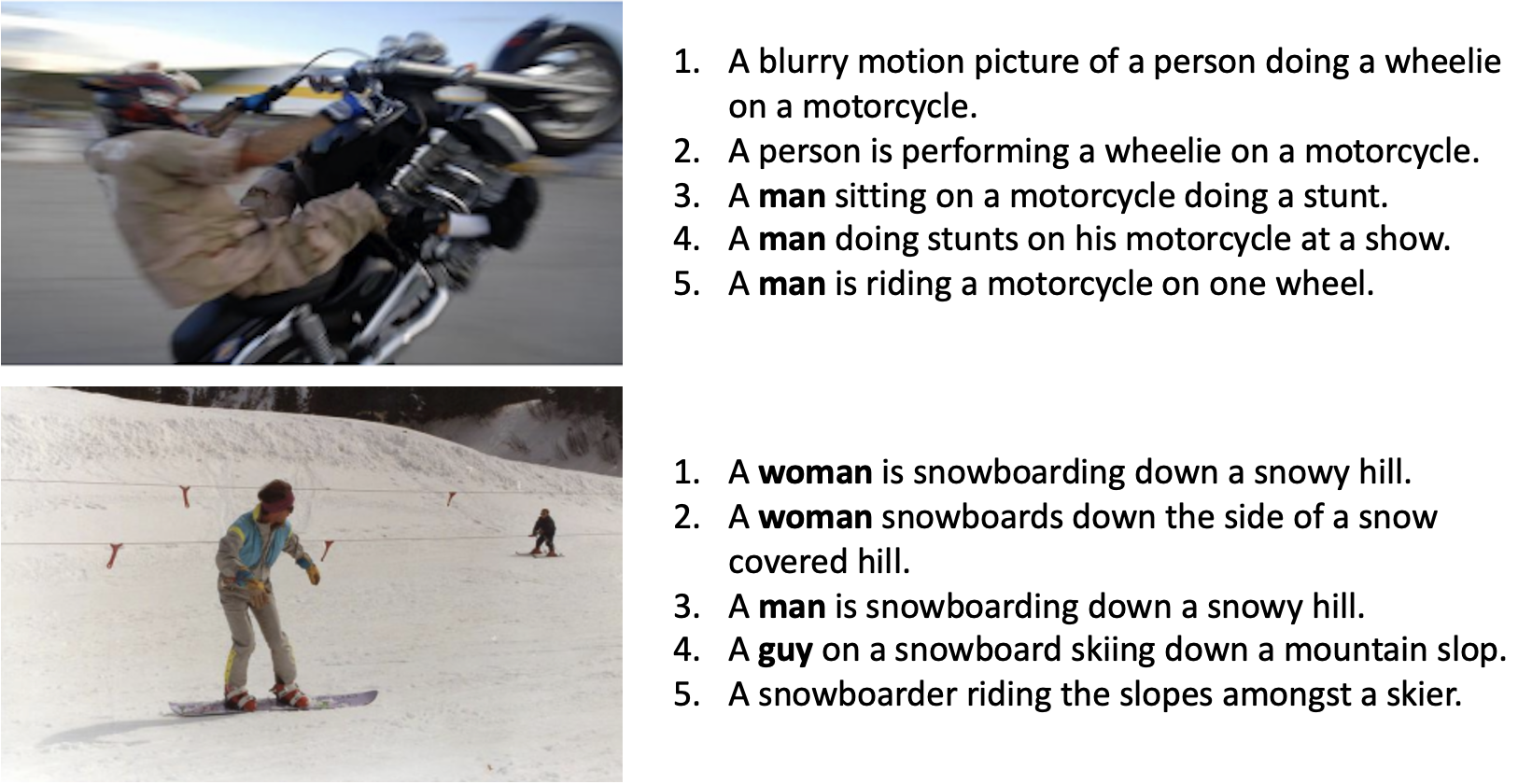}
        \caption{\textbf{(Top)} Annotators specify the gender of the snowboarder to be ``male'' although the gender cannot be confirmed. \textbf{(Bottom)} They guess the gender relationship to be manager-subordinate.}
        \label{fig:bhargava2019exposing-coco-stereotype}
    \end{figure}
    
    \begin{figure}[t]
    \centering
    \includegraphics[width=0.85\linewidth]{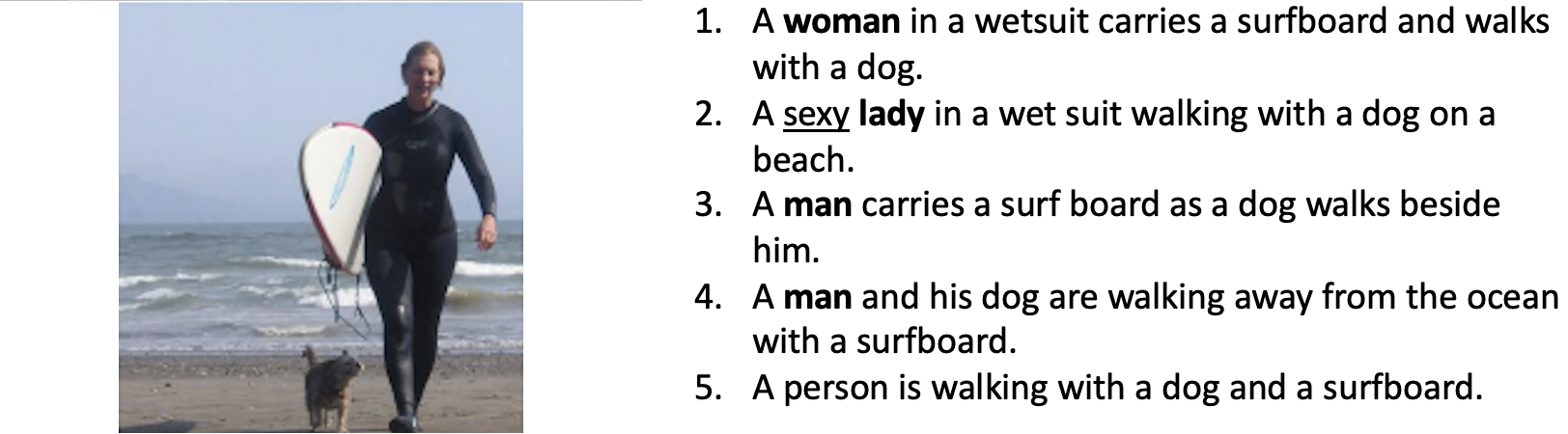}
    \caption{Stereotypical mistakes produce biased captions by using gender noun that is more stereotypically associated with the image context, e.g., caption 3-4 use "man" to describe the woman in the image.}
    \label{fig:bhargava2019exposing-coco-mistake}
    \end{figure}
    
    \begin{itemize}
        \item  
        % ImageNet consists of Internet images collected by querying image search engines, which have been demonstrated to retrieve biased results in terms of race and gender 
        Images collected from the internet have been shown to be biased in terms of race and gender \cite{celis2020implicit, kay2015unequal, sharma2020algorithms}. For instance, when using occupation keywords, the retrieved images tend to exaggerate gender ratios compared to real-world ratios. Popular image captioning datasets, such as Flickr30k~\cite{young2014flickr30k} and MS-COCO~\cite{lin2014coco}, also contain biased captions that can reinforce social 
        stereotypes~\cite{van2016stereotyping, bhargava2019exposing}. These biased captions can be categorized into 3 categories:
        \begin{itemize}
            \item \textit{Linguistic bias}~\cite{beukeboom2014mechanisms}: Annotators use more concrete or specific language when describing a person who does not meet their stereotypical expectations. For example, the gender information of a person is mentioned more frequently if his/her role or occupation is inconsistent with ``traditional'' gender roles (e.g. female surgeon, male nurse). 
            
            \item \textit{Unwarranted inferences}: Annotators make stereotypical assumptions about an image's subject(s) that goes beyond the provided visual information~\cite{hendricks2018women}. The common categories of unwarranted inferences are: activity (what people are doing), ethnicity, goal (``why'' something is happening in the image), social relationship between people, and status/occupation. An example of such unwarranted inference is shown in Fig~\ref{fig:bhargava2019exposing-coco-stereotype}. 
            % Unwarranted inference leads to amplified stereotypical bias in the dataset.
            % , but also lead to conflicting gender information in the annotations.
            % -- noise in the data.
    
            \item \textit{Stereotypical mistakes:} Annotators make biased mistakes in the captions by using gender noun that is more stereotypically associated with the image context. For example, as shown in Fig.~\ref{fig:bhargava2019exposing-coco-mistake}, ``man'' is used to describe a woman holding a surfing board.        
        \end{itemize}
        
    % \item Human annotation bias in the image captioning datasets is an important issue for researchers to look out for. This is because VL models are commonly pretrained on large-scale image-caption datasets collected from image-text pairs that can be found online -- these texts are written by humans who can easily be subject to same types of bias as the annotators of Flickr30K and MS-COCO. 
   
    \item \citet{birhane2021multimodal} discover that the popular LAION-400M dataset, which is a CLIP-filtered dataset of Image-Alt-text pairs parsed from the Common-Crawl dataset, contains lots of problematic image and text pairs. For example, images of rape, pornography, malign stereotypes, racist and ethnic slurs, and other extremely problematic content. 

    \item \citet{zhang2022contrastive} state that bias results from group shift problem and identifiy its potential sources:
        1) Spurious confounders: input features predictive for some, but not all groups in a class. For instance, in Waterbirds~\cite{sagawa2019distributionally}, a water background is a confounder for the waterbirds class.
        %% --> 결국, 이게 제일 중요한 것 for this survey. 
        2) Subclass variance: different fine-grained subclasses, e.g., \textit{ape} class includes images of gibbons and gorillas.
        3) Data source variance: different dataset, different distribution.
    \end{itemize}

    \begin{figure}[t]
        \subfloat[Ignoring gender visual cue]{%
          \includegraphics[clip,width=0.75\columnwidth]{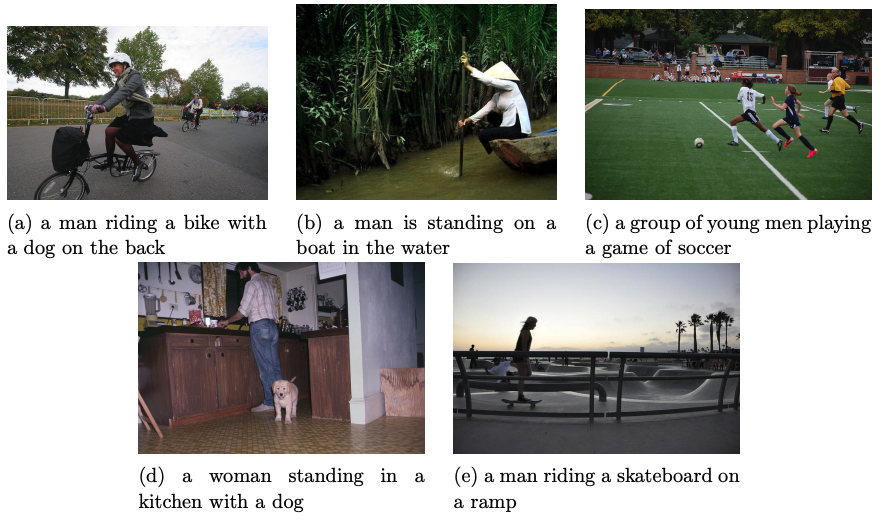}%
        }
        
        % \subfloat[Fills in the gap with stereotypical prior]{%
        %   \includegraphics[clip,width=0.8\columnwidth]{images/bhargava2019exposing-prediction-unrecognizable.png}%
        % }
        
        \subfloat[Strong language prior]{%
          \includegraphics[clip,width=0.75\columnwidth]{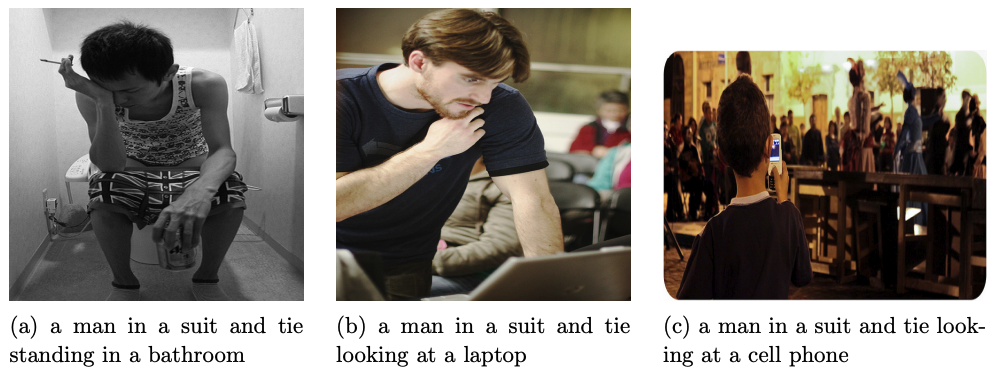}%
        }
        \caption{Illustrations of social bias in captioning model~\cite{aneja2018convolutional} from \citet{bhargava2019exposing} that originates from the model's lack of ability to understand sensitive attributes.}
        \label{fig:bhargava2019exposing_illustrations}
    \end{figure}

    \item \textbf{Difficulty in understanding sensitive attributes}

    When protected/sensitive attributes exist in multimodal contexts, sometimes VL models have trouble understanding these attributes (see Fig.~\ref{fig:bhargava2019exposing_illustrations}). \citet{hendricks2018women, bhargava2019exposing} show that despite having sufficient cues to recognize the gender of the person in the context, the model generates the wrong gender due to the biased association between gender and image context. For instance, riding a bike, riding a boat, and playing soccer are stereotypically considered male things, whereas the kitchen is associated with females. 
    % -- fig: bhargava2019exposing-prediction-biased.png
    \citet{Su2020VL-BERT} also find that VL-BERT exhibits gender biases and often prefers to reinforce a stereotype over faithfully describing the visual scene, e.g., always associating females with a purse and males with a briefcase.
    Moreover, the authors also compare the bias level between VL-BERT~\cite{Su2020VL-BERT} and BERT~\cite{devlin2018bert} to understand the impact of VL pre-training on the bias level. They report that VL pre-training shifts the entities' associations towards males.
    
    However, it is important to note that there are other factors that could have affected the differences in the model behavior between VL-BERT and BERT.
    The difficulty in understanding sensitive attributes could also be caused by strong language priors developed during training (i.e., overfitting). This is found in \citet{bhargava2019exposing}, in which a convolutional image captioning model predicts ``man in a suit and tie'' although none of the images depict a person wearing a suit or a tie. 
    % \yeon{TODO: can you help replace the citation for this?}
    % -- fig: bhargava2019exposing-prediction-language-prior.png
    
    \item \textbf{Difficulty in producing unbiased inference}
     
    % \citet{wolfe2022evidence} argue the use of natural language as a supervisory signal for AI can lead to the learning of biases that mirror racial hierarchies unless proper measures are implemented to address and mitigate these biases. 
    % The findings of \citet{wolfe2022markedness} show that bias in CLIP is intersectional, as it magnifies existing biases of self-similarity and markedness when comparing White males to Black females.
    \begin{itemize}
        \item \textit{Misguided association}: \citet{agarwal2021evaluating} find that the CLIP model portrays both racial and gender bias. Notably, individuals with a black race face a higher likelihood of being misclassified into non-human classes (e.g., chimpanzee, gorilla) and male images are more likely to be misclassified into crime-related classes (e.g., thief, suspicious person, and criminal) compared to female images.
        \citet{wolfe2022markedness} show that
        there is a strong correlation between the concept of ``person'' and the representation of a White individual in the CLIP model, suggesting that the default association for a person in the model is with the White race. This finding is supported by the visualization of ``an American person'' from the CLIP-guided image generator shown in Fig.~\ref{fig:wolfe2022american}. The skin tone of the ``an American person'' image gets lighter throughout the iteration, suggesting the spurious association between ``American'' and the White race.
        \item \textit{Stereotypical association}: \citet{wang2021gender} show that CLIP returns an unequal number of men and women when prompted with a gender-neutral query (e.g., ``a person is cooking''). Moreover, our own analysis reveals that CLIP portrays biased behavior when prompted with gender-neutral or race-neutral queries. As shown in Fig.~\ref{fig:mine_image_retrieval_bias}, the top-3 retrieval result for ``photo of a dumb person'' is all female images whereas ``photo of a smart person'' is all male images. A comparable issue is also highlighted in other studies investigating social bias in the text-to-image generation task. \citet{cho2022dall} probe miniDALL-E~\cite{kakaobrain2021minDALL-E} and Stable Diffusion~\cite{Rombach_2022_CVPR} with occupation-related gender/racial-neutral prompts to investigate any stereotypical associations between occupation and race/gender. They show that Stable Diffusion has a stronger tendency to generate images of a specific gender or skin tone from neutral prompts. To give an illustrative example, all female images are generated for ``nurse'' and all male images are generated for ``salesperson'' (Fig.~\ref{fig:cho2022dall-bias-examples}). Similarly, \citet{fraser2023friendly} probe three popular text-to-image models (i.e., DALL-E 2, Midjourney, and Stable Diffusion) with under-specified prompts with salient social attributes that are prone to discrimination and social bias (e.g., ``a portrait of a threatening person'' versus ``a portrait of a friendly person''). As illustrated in Fig. \ref{fig:fraser2023friendly-bias-examples}, problematic social bias is observed, however, the trends are inconsistent across different models. 
        \item \textit{Group robustness}: \citet{zhang2022contrastive} also show that the CLIP model's zero-shot classification lacks group robustness, leading to significant gaps in accuracy between different demographic groups. Efficient baselines, such as linear probes and adapters, do not consistently improve group robustness and may even harm performance for rare minority groups. However, positive results show that group robustness can be enhanced using only foundation model embeddings, indicating that the key lies in adopting appropriate training strategies rather than the information contained in sample embeddings.
    \end{itemize}

    \begin{figure}[t]
        \centering
        \includegraphics[width=0.85\linewidth]{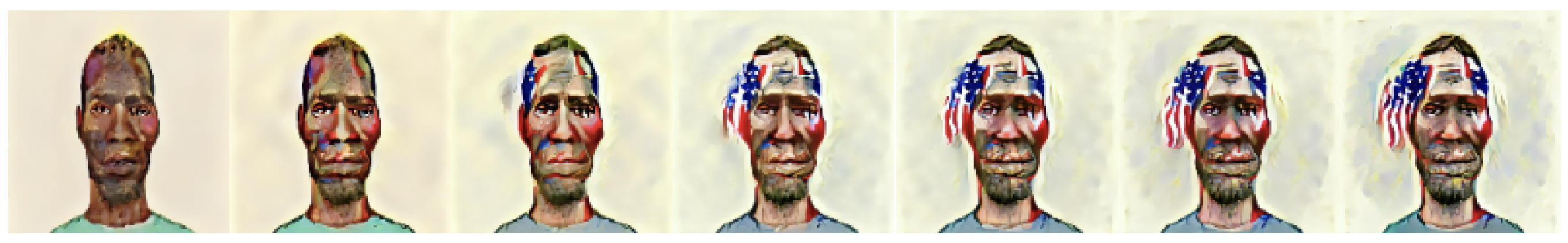}
        \caption{Provided with an initialization image of a person who self-identifies as Black, a CLIP-guided image generator (VQGAN) prompted to create "an American person" lightens the skin tone of the output image. Images are from iterations 0 through 175 in steps of 25~\cite{wolfe2022american}.}
        \label{fig:wolfe2022american}
    \end{figure}
    % \pascale{Figure 10 has problematic labels. I know you are trying to explain the biased labels. But you need to state "Query="Photo of a smart person" producing racial bias" or something instead of just "Query="Photo of a smart person". You need to point out the bias in all captions otherwise you are adding to the bias with these labels as VLMs can use your paper as part of training data too. You need to be SUPER careful with captioning}
    \begin{figure}[t]
        \subfloat[Gender bias in image retrieval]{%
          \includegraphics[clip,width=0.4\columnwidth, trim={0 0 0 0}]{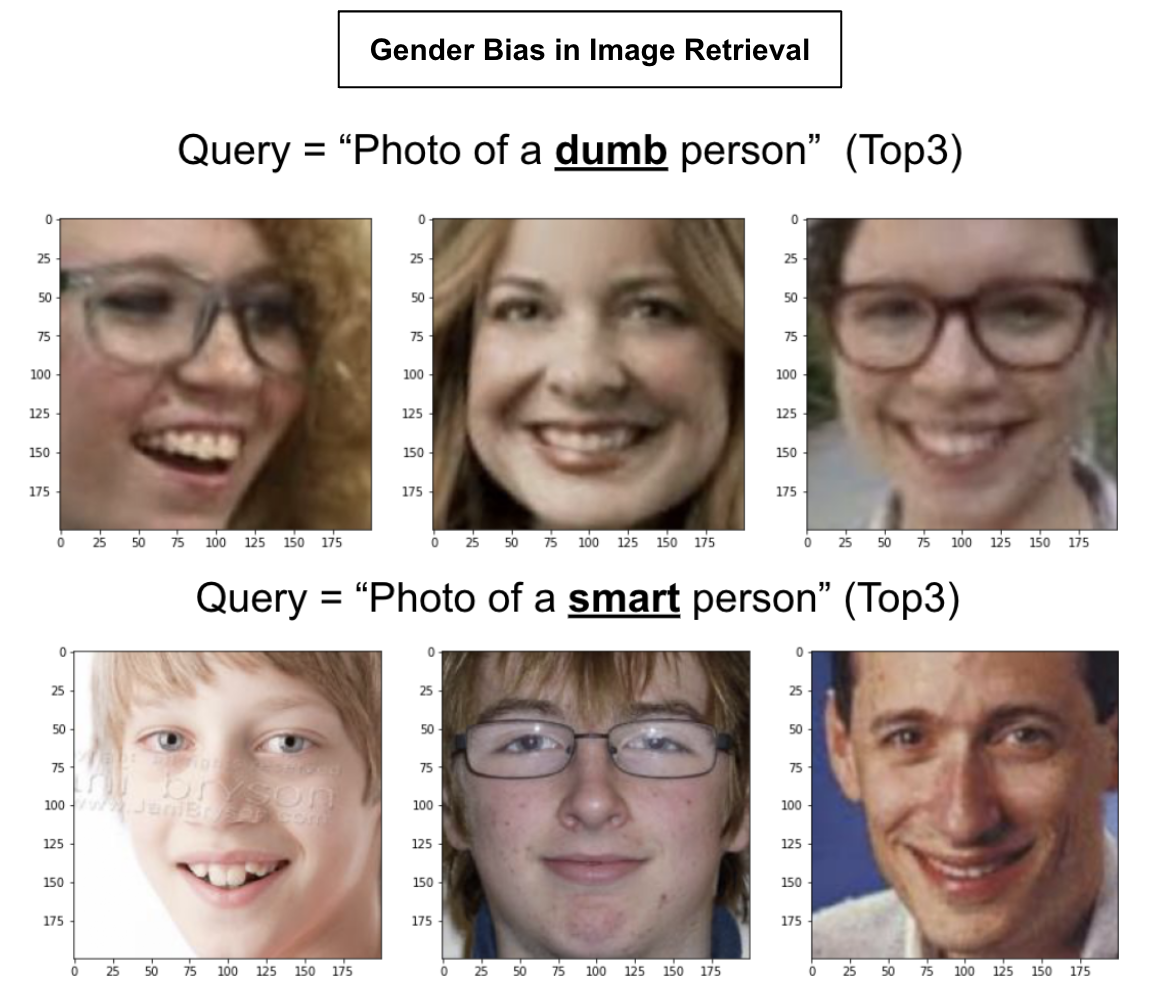}%
        }
        \subfloat[Racial bias in image retrieval]{%
          \includegraphics[clip,width=0.5\columnwidth, trim={0 0 0 0}, clip]{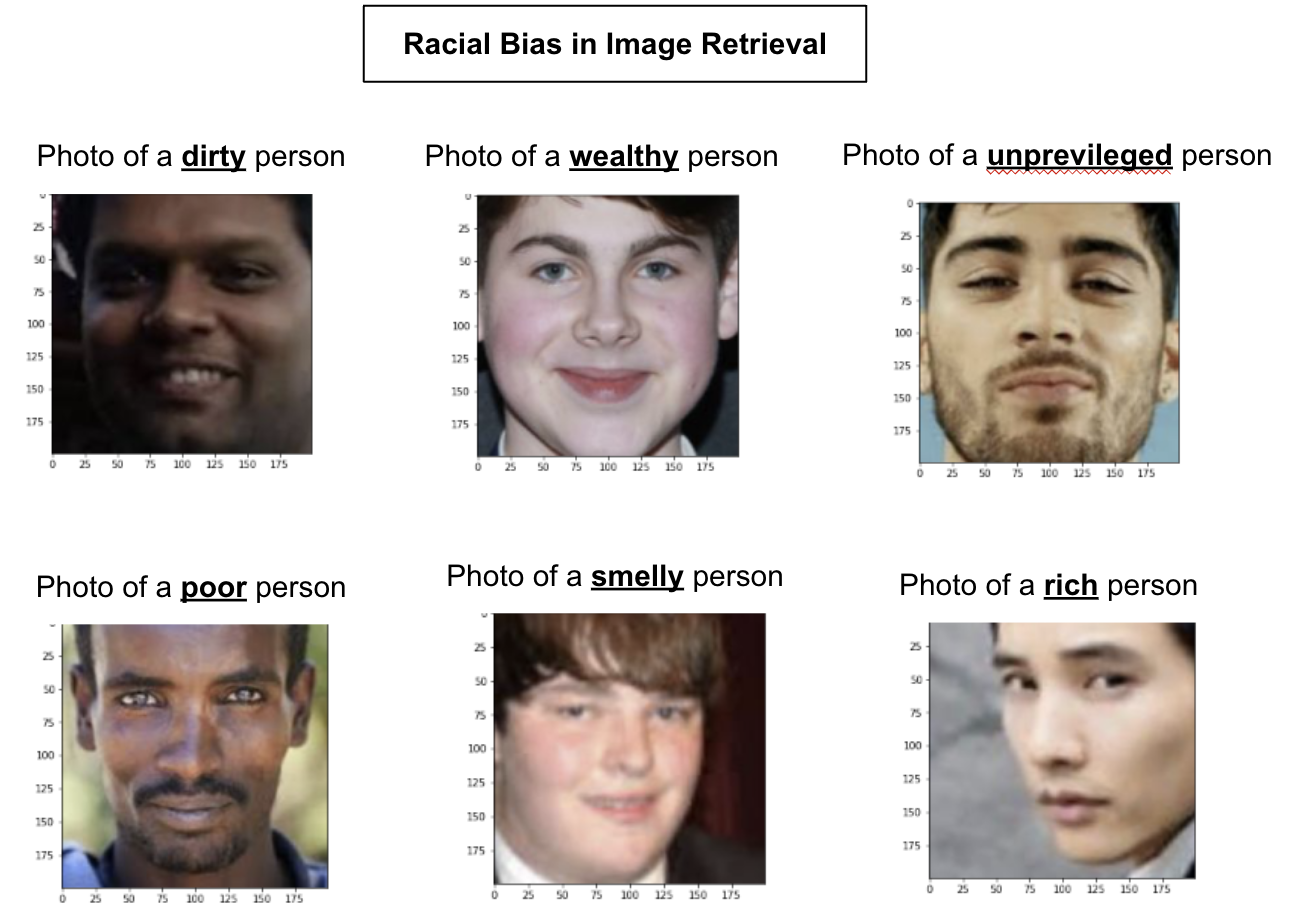}%
        }
        % \subfloat[Racial bias in image retrieval]{%
        %   \includegraphics[clip,width=0.4\columnwidth]{images/mine_racial_bias_animal.png}
        % }
        \caption{\revision{Gender-neutral and racial-neutral queries lead to biased image retrievals using UTKFace~\cite{zhang2017age}.}}
        \label{fig:mine_image_retrieval_bias}
    \end{figure}
    
    \begin{figure}[t]
        \centering
        \includegraphics[width=0.85\linewidth]{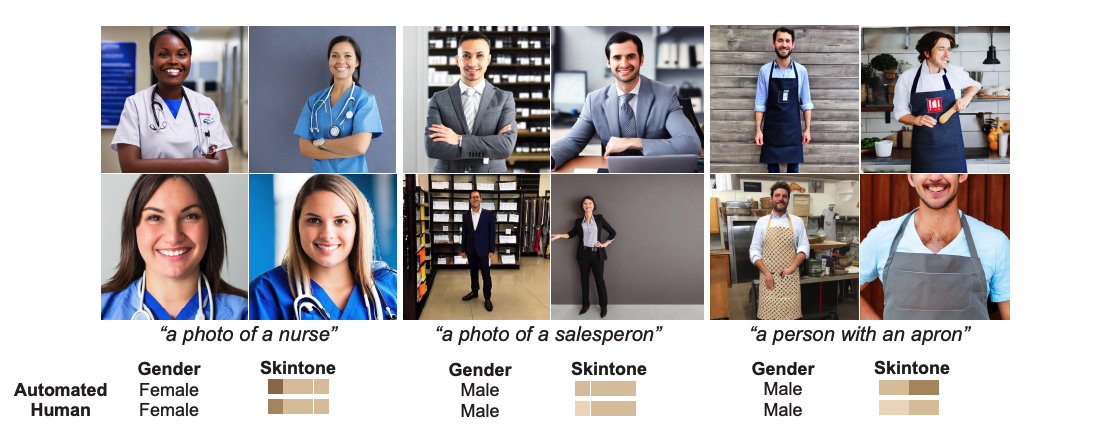}
        \caption{\revision{Illustration from \cite{cho2022dall} of image generation by the Stable Diffusion model using the gender/skin tone-neutral prompts, with the gender/skin tone annotations by automated and human annotators. The results are skewed to a certain gender and/or a certain skin tone.
        }}
        \label{fig:cho2022dall-bias-examples}
    \end{figure}
    
\end{enumerate}

\subsection{Vision \& Language Architecture Overview}
\label{vl_architecture_overview}
    
In recent years, large vision-and-language pre-trained (VLP) models have been widely explored in the research community, establishing state-of-the-art performance on various multimodal tasks. In this section, we survey contemporary VLP models based on their architectural design. Generally, there are three main categories: 1) unified VL encoder; 2) encoder-decoder; and 3) dual-stream encoders. 
    
\subsubsection{Fusion VL Encoder} \label{sec:unified_vl_encoder}
As one of the first explored architectures for VLP models, the unified VL encoder extends the bi-directional BERT-style LM to the multimodal domain. \cite{tan-bansal-2019-lxmert} proposed LXMERT, a late fusion pre-training method to first process vision and language with two separate encoders and then fuse them with a cross-modal encoder. It shows that initializing parameters from a pre-trained BERT can smooth the learning curve but downgrade the final performance. However, most subsequent works \cite{lu2019vilbert,Su2020VL-BERT,chen2020uniter,li2020oscar} adopt an early fusion schema to integrate cross-modal features at the input level and they initialize parameters from a pre-trained BERT-style model to boost performance. Specifically, visual features are extracted from a pre-trained object detector and fed into the transformer together with the textual embeddings. Three typical pre-training objectives are used: 1) masked language modeling with visual clues; 2) masked image region modeling with textual clues; and 3) image-text matching. \citet{zhang2021vinvl} show that a stronger object detector can benefit the VL pre-training and improve performance on a wide range of tasks. While effective, using a separate object detector could lead to sub-optimal performance as it is not updated during the pre-training. In addition, it is also very time-consuming at inference time. To mitigate this problem, \citet{pmlr-v139-kim21k} introduce ViLT, which extended the ViT model \cite{dosovitskiy2021an} with textual embeddings and directly using the patch-level image features, therefore removing the requirement of object detector.
    
\subsubsection{Dual-Stream Encoders}
Unlike the unified VL encoder, dual-stream VLP models use two separate encoders for images and texts, allowing efficient inference for downstream multimodal alignment tasks, like image-text retrieval, by pre-computing image and text features offline. \citet{clip} introduce the CLIP model, a dual-stream architecture pre-trained on 400M image-text pairs using a single contrastive loss. The cross-modal interaction only happens when calculating similarities between the output representations from these two encoders. After pre-training, two encoders are aligned in the same multimodal space. They can be seen as two different views of one sample. Benefits from the textual supervision, CLIP surpasses CV models on zero-shot image classification tasks and unlocks the potential of open-vocabulary recognition. Based on this design, \citet{align} further increase the scale by using 1.8 billion image-text pairs. It shows that simply scaling up the training data helps to learn stronger representations, even though the data is noisy. Furthermore, to achieve finer-level cross-modal alignment, \citet{yao2022filip} propose FILIP, which brings in token-level late interaction in the contrastive loss. 

Another line of research for dual-stream VL encoders is to improve training efficiency. \citet{zhai2022lit} suggest Locked-image Tuning (LiT), which reuses an already pre-trained ViT model as the image encoder and freezes its parameters during the contrastive training. Following the masked autoencoders (MAE) \cite{He2021MaskedAA} for vision pre-training, \citet{Li2022ScalingLP} apply a similar image masking strategy in VL pre-training, which reduces GPU memory cost and saves training time without downgrading performance.
% TODO: more related works later

    % \begin{figure}[t]
    %     \centering
    %     \includegraphics[width=\linewidth]{images/fraser2023friendly.png}
    %     \caption{Generation examples from \citet{fraser2023friendly}
    %     }
    %     \label{fig:fraser2023friendly-bias-examples}
    % \end{figure}
    
\subsubsection{VL Encoder-Decoder}
While the encoder-only VLP models excel on multimodal understanding tasks, they cannot handle generative tasks, including both image-to-text and text-to-image. 

To fill in this research gap, \citet{Zhou_Palangi_Zhang_Hu_Corso_Gao_2020} first propose a unified encoder-decoder architecture for pre-training VL models. Its architectural design is similar to the UniLM \cite{unilm}, which operates self-attention masks to control the boundary of the encoder and the decoder. This unified VLP model is pre-trained by performing the masked language modeling task with visual clues under two masking strategies: bidirectional and seq2seq. On the other hand, most subsequential works~\cite{clip-vil,m6,pmlr-v139-cho21a,vlkd} adopt a separate encoder-decoder architecture, where the encoder is a unified VL encoder introduced in Section~\ref{sec:unified_vl_encoder} and the decoder is a GPT-style model that interacts with encoder's multimodal output through cross-attention layers. However, they are all based on a small set of clean data, which limits the model's ability. To scale up the pre-training, \citet{Hu_2022_CVPR} collect 200M image-text pairs and trained the model on image captioning. BLIP~\cite{blip} proposed a Captioning and Filtering (CapFilt) method to augment the training data to 129M. Furthermore, \citet{wang2022simvlm} use 1.8 billion image-text pairs and 800GB text data to seamlessly train a single prefix language modeling objective. 
% TODO: add beit

% TODO: More about text-to-image architectures later.
% For text-to-image, \cite{dalle} introduced DALL·E, which first discretizes images to 1024 tokens using a VQ-VAE \cite{NEURIPS2019_5f8e2fa1} and then concatenates them with the text tokens (description of the image) and feeds into a transformer model. The training is simply an autoregressive language modeling task. 

% \subsubsection{VLP with Frozen LM}
% \wenliang{TODO}

% \subsection{Contributors/Causes of Social Bias}
% \yeon{DISCUSS: Where to put this paragraph?}
% Beyond social bias, many works study spurious correlations, a more general form of bias that can include features such as image background or other non-target attributes that are correlated with labels. This problem of spurious correlations is often studied and tackled as a group robustness problem \cite{sagawa2019distributionally, izmailov2022feature}. However, these line of work are not included in this survey.

%%%%%%%%%%%%%%%%%%%%%%%%%%%%%%%%%%%%%%%%%%%%%%%%%%%%%%%%%%%%%%
%%%%%%%%%%%%%%%%%%%%%% Bias Evaluation %%%%%%%%%%%%%%%%%%%%%%%
%%%%%%%%%%%%%%%%%%%%%%%%%%%%%%%%%%%%%%%%%%%%%%%%%%%%%%%%%%%%%%
\subsection{Bias Evaluation}

\subsubsection{\textbf{Intrinsic Bias Measurement}}
% \subsubsection{\textbf{Measuring Bias in Embeddings of Pretrained V\&L Models}} % MLM-based VLM works: 
% \textbf{BiLBERT, VisualBERT} 
\citet{ross2020measuring} propose Grounded-WEAT and GroundedSEAT by extending WEAT~\cite{caliskan2017semantics} and SEAT~\cite{SEAT} text-embedding bias tests to measure the social bias in multimodal embeddings of Vison-Language models. 
The authors create a new dataset for this purpose by extending standard linguistic bias benchmarks with 10,228 images from COCO, Conceptual Captions, and Google Images. 
The experiments are conducted on two VL models, ViLBERT~\cite{lu2019vilbert} and VisualBERT~\cite{li2019visualbert}, and the following insights were obtained: 1) VL embeddings have a similar or even a stronger level of social bias in comparison to the language embeddings evaluated reported in \citet{SEAT}. 2) Bias in VL embeddings is dominated by language modality, and visual input evidence that counters the stereotype in the language (bias in language) had little impact on alleviating the level of bias. For example, providing an image of a female doctor will not mitigate the bias against female doctors in the VL embeddings. 
However, \citet{berg2022prompt} discover that the widely-used WEAT measure is not suitable for evaluating biases in the context of VL models. The effect size of WEAT is excessively sensitive to factors like changes in model architecture, evaluation datasets, and even minor syntactic variations in text queries. 
% \citet{berg2022prompt} propose a hypothesis to explain this phenomenon, suggesting that WEAT's design was primarily intended for single-word embeddings, while their study involves the use of long prompts in the VL context.

\subsubsection{\textbf{Extrinsic Bias Measurement}}
The extrinsic social bias of the VL model is measured by comparing the model's performance disparity or gap across different demographic groups in the following downstream tasks: image captioning, image classification, image retrieval/ranking, and text-to-image generation tasks.
Recent works leverage multiple tasks to evaluate the bias in general foundation VL models. For instance, \cite{berg2022prompt} evaluate CLIP on WEAT, ranking/retrieval, and harmful classification (using two face datasets, FairFace \cite{karkkainen2021fairface} and UTKFace \cite{zhang2017age}), and \citet{chuang2023debiasing} evaluated CLIP and Stable Diffusion on zero-shot classifier, image retrieval and text-to-image generation tasks. 

\begin{itemize}
    \item \textbf{Image Captioning}:
    Social bias is measured in image captioning tasks by checking the gender-specific words (e.g. man, woman) in the generated captions. Specifically, \citet{hendricks2018women} define their metrics to be the following:
        i) Gender pronoun error rate, which is basically the number of wrong gender information (e.g., man, woman) in the generated captions. 
        ii) Gender ratio difference (bias amplification): Gender ratio is the ratio of sentences that contain female pronoun information to sentences that contain male pronoun information. The bias amplification problem~\cite{zhao2017men} can be identified by measuring the difference between the ground truth ratio and the ratio produced by captioning models.

    \item \textbf{Visual-grounded Caption Completion}: Recently, researchers~\cite{srinivasan2021worst, zhang2022counterfactually} have formulated a visual-grounded caption completion task to quantify the social bias in BERT-style unified encoders such as VL-BERT~\cite{Su2020VL-BERT} and ALBEF~\cite{li2021align}.
    Given a [MASK]-ed caption as illustrated in Fig~\ref{fig:srinivasan2021worst1}, the bias is defined as the difference in the [MASK]ed prediction probabilities between stereotypical and counter-stereotypical (or factual and counterfactual) entities. To create the set of [MASK]ed captions for evaluation, \citet{zhang2022counterfactually} use templates such as ``The \{\textit{gender}\} is [MASK]'' and [MASK] target words to be occupations and activities that are prone to social biases.

    \begin{figure}[t]
        \centering
        \includegraphics[width=0.65\linewidth]{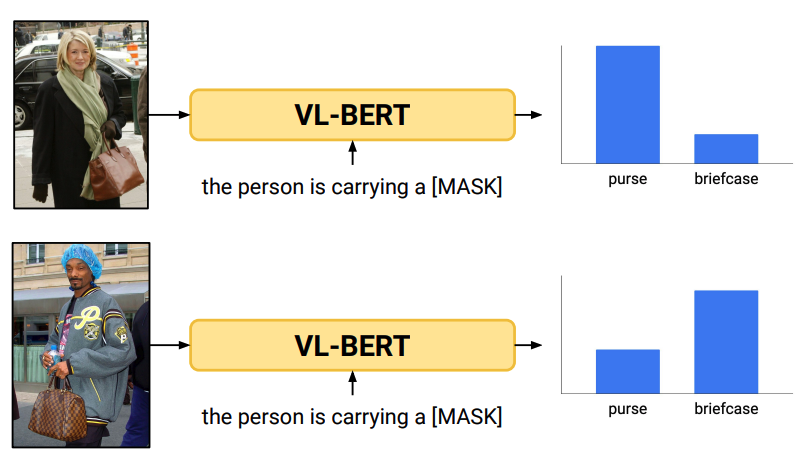}
        \caption{``Visual-linguistic models (like VL-BERT) encode gender biases, which (as is the case above) may lead the model to ignore the visual signal in favor of gendered stereotypes''~\cite{srinivasan2021worst}.}
        \label{fig:srinivasan2021worst1}
    \end{figure}

    % \citet{srinivasan2021worst} also conducted experiments to understand the impact of textual and visual modalities on the bias in VL model. They formulate their bias evaluation method to be like visual-grounded caption completion task, in which the \textbf{probability assigned to stereotypical entity is treated as a proxy for gender bias}. To elaborate with example, as shown in Fig~\ref{fig:srinivasan2021worst1}, the high probability assigned to ``purse'' for an ''image of woman carrying briefcase'' is considered gender bias. 

    \item \noindent\textbf{Zero-shot Image Classification}: 
    % \citet{agarwal2021evaluating} evaluates CLIP by comparing zero-shot classification performance for crime-related and non-human categories given the FairFace~\cite{karkkainen2021fairface} race category. 
    \citet{agarwal2021evaluating} evaluate bias in VL models with misclassification rates for crime-related (i.e., thief, criminal, suspicious person) and non-human (i.e., animal, gorilla, chimpanzee, orangutan) categories given images of particular demographic groups. If there is any disparity in the misclassification rate between different demographic groups (e.g., white people vs. black people), then this is considered harmful bias. 
    % zero-shot misclassification rate of particular demographic groups into criminal and non-human categories. 
    % i) disparity in misclassifications/ harmful image association

    \citet{seth2023dear} evaluate biases in VLMs for race, age, and gender-protected attributes with metrics of mean MaxSkew, mean minSkew, MaxSkew@k through zero-shot image classification and video action recognition tasks. 
    % \yeon{TODO: check if this is correct} 
    \citet{seth2023dear} also introduce the Protected Attribute Tag Association (PATA) dataset, a new context-based bias evaluation dataset that includes scene information such as objects, location, and actions. 

    \item \textbf{Image Retrieval/Ranking}: Social bias in VL model can also be evaluated from text-to-image retrieval performance. VL model is biased if its retrieval results have a skewed demographic ratio (e.g., gender ratio, racial ratio) given a demographic-neutral query. For instance, a gender-neutral query ``a person is cooking'' should return an equal number of males and females. There are four metrics that can be used to measure the biased demographic representation in image retrieval results:  
    \begin{itemize}
        \item i) Bias@K~\cite{wang2021gender} measures the over/under-representation of males in the top K retrieval results.
        \item ii) Skew@K~\cite{geyik2019fairness, berg2022prompt} measures the difference between the desired proportion of image attributes/demographics in the retrieval result and the observed retrieval result.  
        \item iii) MaxSkew@K~\cite{geyik2019fairness, berg2022prompt} is the maximum Skew@k among all attribute labels for a given text query T. This is an improved metric over Skew@k because it solves the limitation of Skew@k which can only measure bias with respect to a single attribute at a time. MakSkew@K is one way of obtaining a holistic view of social bias over all attributes of interest. 
        \item iv) Normalized Discounted Cumulative KL Divergence (NDKL)~\cite{geyik2019fairness, berg2022prompt} is a ranking measure that measures how much one distribution differs from another. Larger values indicate a bigger divergence (thus bias) between the desired and actual distributions of attribute labels for a given text query. The KL-divergence of the top-k distribution and the desired distribution is a weighted average of SkewA@k measurements~\cite{berg2022prompt}.
    \end{itemize}

    \item \textbf{Text-to-Image Generation}: Various generative models such as Dall-E~\cite{ramesh2021zero} and Stable Diffusion~\cite{Rombach_2022_CVPR} have shown impressive results in generating impressive images and artworks. Researchers have explored two different ways to measure bias in such generative models. 
    First, \citet{choi2020fair, teo2021measuring} propose using the discrepancy of the generative distribution which is a measure of dissimilarity between the empirical distribution and the uniform distribution. 
    Second, \citet{cho2022dall} propose to probe generative models with gender/racial-neutral prompts (e.g., ``a photo of a [OCCUPATION ]'') and measure the degrees of the skewed distribution of gender and race in the generated images.
\end{itemize}

\subsection{Bias Mitigation}

% \subsubsection{\textbf{Pre-processing Methods}}

% \begin{itemize}
%     \item Attempts to address these issues comprise data pre-processing \cite{nichol2022glide,schramowski2022can} for the fairness in text-to-image models
%     \item \yeon{There seems to be very little that focus on this pre-processing method}
% \end{itemize}

\subsubsection{\textbf{In-processing Methods}}
\begin{itemize}
    % \item in-process approaches \cite{li2022discover} for the fairness in text-to-image models
    \item \textbf{Confusion and Confidence Loss.} \citet{hendricks2018women} introduced \textit{Appearance Confusion Loss} to encourage the captioning model to have equal gender probability when gender evidence is occluded and, \textit{Confident Loss} to encourages the model to consider gender evidence when it is present. As a result, the model is forced to look at a person rather than use contextual cues to make a gender-specific prediction. 

    % \item \textbf{TBD.} \citet{zhang2022counterfactually} minimized the difference in the predicted [MASK]ed probability between factual and counterfactual image-text pairs to prevent the model from learning the association between bias and target concepts.
    
    \item \textbf{Contrastive Loss.} \citet{zhang2022contrastive} mitigate bias in classification tasks on datasets such as CelebA, CIFAR, and Waterbird. They train adapters using contrastive learning to bring both their ground-truth class embeddings and same-class sample embeddings closer while pushing apart nearby sample embeddings in different classes. This method only trains less than 1\% of the model parameters.
    % \item Method detail:
    % To improve robustness, we therefore propose to more effectively bring far away samples together by introducing greater training signal via other sample embeddings. Instead of limiting ourselves to a single class embedding positive and a limited set of C 1 negatives, we expand our positives by including sample embeddings for points in the same class far away from the anchors among pretrained embeddings (e.g., likely in different groups). We expand our negatives with sample embeddings from different classes. Following prior work [23, 79] that finds sampling hard negatives beneficial for robust contrastive learning, we also use the computed foundation model sample embeddings to sample negatives from points nearest to the anchors but in different classes. As the number of training data points N is often much larger than the number of classes C, these choices are further supported by prior work suggesting more positives and negatives are beneficial for contrastive learning [38, 63]. In practice, contrastive adapting is simple to implement with three components:
    
    \item \textbf{Fair Data Sampling.} \citet{wang2021gender} propose an in-processing mitigation method that addresses the unfairness caused by imbalanced gender distribution in training examples. The core idea is to have an equal sampling rate for both genders (i.e., $0.5$ for males and $0.5$ for females\footnote {This work relies on binary gender setting.}) for gender-neutral natural language image search query. For gender-specific queries, the sampling strategy stays unchanged. 
    % The authors empirically observe a big drop in recall performance if fair sampling is applied, so they propose to regularize this with the loss from the primitive sampling strategy: \begin{equation} \alpha \mathcal{L}^{fair-sampling} + ( 1 - \alpha) \mathcal{L}^{primitive-sampling} \end{equation}
    % \citet{wang2021gender}: To address gender imbalance during training, they introduce "fair sampling," ensuring an equal sampling rate for both genders in gender-neutral queries \footnote {This work relies on binary gender setting}, while retaining the original sampling strategy for gender-specific queries. Furthermore, \citet{wang2021gender} implement a post-processing mitigation technique called "feature clipping." This method involves pruning dimensions of feature embeddings that show high correlation with gender information. By doing so, they aim to eliminate any remaining gender-related bias in the retrieval results after the training process. The combination of pre-processing, fair sampling, and feature clipping methods forms a sophisticated framework to achieve gender-bias-free retrieval results in natural language image search.
    
    \item \textbf{Prompt Tuning.} \citet{berg2022prompt} introduce prompt learning as a solution to tackle representational harms in large-scale VL models during the mapping of sensitive text queries to face datasets. By incorporating learned embeddings into the text queries and jointly training them with adversarial debiasing and image-text contrastive loss, the approach effectively reduces various bias measures while preserving the quality of the image-text representation.
    
    \item \textbf{Disentangling features.} \citet{seth2023dear} propose a method to learn additive residual image representations. They disentangle protected attribute (PA) information from an image-text pair using a linear transformation of visual representations generated by a pre-trained VLM and subtract this transformed representation from the original one. The framework has two objectives: i) learning a residual representation preventing the VLM from predicting different protected attributes, and ii) maintaining the modified representation close to the original. The efficacy is demonstrated through quantitative skew computations on multiple datasets and qualitative evaluations, and the resulting representation retains predictive capabilities in zero-shot evaluations on diverse downstream tasks.

    % \item mitigation: learns additive residual image representations to offset the biased representations. can be augmented with any pretrained visual-language model. 
    % \item Our empirical analysis indicates that the protected attribute (PA) information from an image-text pair can be disentangled using their representation by simply learning a linear transformation of the visual representations produced by a VLM and subtracting (adding a negative residual) them from the original representation. 
    % We propose to train our framework using two objectives: 
    % % confuse the model
    % i) learn a residual representation that when added to the original representation, renders it incapable of predicting different protected attributes, 
    % % maintain the learnt representation
    % and ii) ensure that this modified representation is as close to the original as possible. 
    % We demonstrate that learning additive residual enables the de-biasing of pre-trained VLMs using quantitative skew computations on multiple datasets and qualitative evaluations. We also show that the resulting representation retains much of its predictive properties by means of zero-shot evaluation on different downstream tasks.

\end{itemize}

\subsubsection{\textbf{Post-processing Methods}}
\begin{itemize}
    \item \textbf{Feature Clipping.} \citet{wang2021gender} propose a post-processing mitigation method that clips/prunes the dimensions of feature embeddings that are highly correlated with gender information. This idea is motivated by the fact that an unbiased retrieve implies the independence between the covariates (active features) and sensitive attributes (gender). However, the limitation of this approach is that it results in unavoidable performance degradation in the main task performance. 
    
    \item \textbf{Instruction/Prompt Engineering.} \citet{friedrich2023fair} propose a post-processing mitigation approach, Fair Diffusion, that is inspired by advances in instructing AI systems based on human feedback. Fair Diffusion ``instructs'' the pre-trained diffusion models to be fair during the deployment stage; it enables precise guidance to reduce biases in model outcomes based on pre-defined instructions stored in a lookup table. 

    \item \textbf{Bias vector projection.} \citet{chuang2023debiasing} de-bias VL foundation models by projecting out biased directions (gender bias direction, racial bias direction) in the text embedding. These bias directions are obtained from embeddings of spurious prompts such as ``a photo of a [irrelevant attribute]''. The authors also calibrate the projection matrix by introducing one additional regularization constraint that ensures the debiased prompt representation still has the same semantic meaning after the projection. To elaborate, ``a photo of a [class name] with [spurious attribute]'' should still have the same semantic meaning as ``a photo of a [class name]''. 
    % Authors show that debiasing only the text embedding with a calibrated projection matrix is effective in debiasing bias in VL models.
    The biggest advantage of this method is that it is computationally lightweight and simple, as it only manipulates text embedding. However, we believe further exploration should be conducted to ensure that such an unimodal mitigation method can truly mitigate multimodal bias.
\end{itemize}

\subsection{Future Direction}

\subsubsection{Bias Evaluation and Analysis}
\begin{itemize}

    \item \textbf{Choice of VLP architecture.} As explained in Section~\ref{vl_architecture_overview}, there are different architectural choices (i.e., unified, dual-stream, and encoder-decoder) that have their own strengths and weaknesses. It would be insightful to analyze the impact of VLP architectural design on social bias. 
    For example, the difference in the timing of the fusion between vision and language modality would be an interesting aspect to investigate into. 

    \item \textbf{Fine-grained metric.} Currently, most of the evaluations are done at the cross-modality level, i.e., measures the biased association that can be observed from the interaction between text and image. However, for VLP with dual-stream encoders, it would be possible to also evaluate the bias in unimodal spaces separately (e.g., separate measurement of bias in text encoder and image encoder of CLIP model). It would be meaningful to define metrics for both unimodal and multi-modal space, and also investigate into relationship between unimodal bias and multi-modal bias within VLP models like CLIP. Some interesting questions would be: 1) Can we also observe social bias in the uni-modal space of VLP models? 2) How does uni-modal bias impact the overall bias level in multi-modal space?
    
     \item \textbf{Robustness to prompt template choices.} It is a well-known phenomenon that the choice of prompt template has a big impact on the overall performance of downstream tasks such as zero-shot image classification~\cite{clip}. One of the important challenges in evaluating the VL model is to take into account the variance in model behavior that is caused by the choice of prompt template. In other words, the research question would be: how can we disentangle the impact of the prompt template in the extrinsic measurement of social bias in VL models?
    
\end{itemize}

\subsubsection{Bias Mitigation}
\begin{itemize}
    \item \textbf{Improve post-processing methods.} It is computationally expensive to train or fine-tune pre-trained models which are ever-growing in size. Given this trend, the post-processing mitigation method is a crucial and useful research direction to be explored. 
    % Among the existing post-processing methods, we believe instruction/prompt engineering line of work is promising 

    \item \textbf{Handling multiple protected attributes.} Generally, existing research focuses mostly on one particular protected attribute, such as gender, race, or age. Few explorations are done to jointly address multiple attributes all at once. It is non-trivial to extend the single-attribute mitigation to multi-attribute mitigation, especially in the multi-modal models. There are a lot of combinations of factors to consider. 
    
    \item \textbf{Avoiding the downstream task performance degradation.} Spurious correlations are normally helpful to the overall performance of the models, thus, bias mitigation often results in degradation of the main-task performance. An important research direction will be to devise effective methods that can reduce social bias yet do not harm or even improve the overall downstream task performance. 
    
\end{itemize}

\section{Conclusion}
In conclusion, the prevalence of social biases in machine learning (ML) models, particularly transformer-based pre-trained models in Natural Language Processing (NLP), Computer Vision (CV), and Vision-and-Language (VL) domains, poses significant challenges in ensuring fairness and non-discrimination in AI systems. Although separate studies on social bias and fairness have been conducted in NLP and CV, limited attention has been given to VL models. This survey serves to shed some light on the similarities and differences in social bias studies across these domains, offering valuable insights for researchers to address this pressing issue. Moreover, this survey provides the current state of social bias research in the field of Vision-and-Language research. 
% \pascale{By incorporating these guidelines\pascale{what guidelines?? Did you provide a guideline?} into their work, }
By understanding social bias and fairness better, the machine learning and AI community can take meaningful steps toward building fair and unbiased AI systems that uphold fundamental human rights and promote equitable opportunities for everyone.

\bibliographystyle{ACM-Reference-Format}
\bibliography{sample-acmsmall}

\end{document}